\documentclass[smallextended]{svjour3}

\usepackage[inline]{enumitem}
\usepackage[normalem]{ulem}
\usepackage[skins,breakable,most]{tcolorbox}
\usepackage{amsmath,amsfonts}
\usepackage{array}
\usepackage[round]{natbib}
\usepackage{hyperref}
\makeatletter
\@ifundefined{c@chapter}{\let\cl@chapter\@empty}{}
\makeatother
\usepackage[capitalize]{cleveref}
\crefname{section}{Sect.}{sections}
\Crefname{section}{Section}{Sections}

\usepackage{framed}
\usepackage{lipsum}
\usepackage{listings}
\usepackage{longtable}
\usepackage{mathtools}
\usepackage{multirow}
\usepackage{pbox}
\usepackage{relsize}
\usepackage{rotating}
\usepackage{tabularx}
\usepackage{booktabs}
\usepackage{todonotes}
\usepackage{url}
\usepackage{subcaption}
\usepackage{graphicx}
\usepackage{xspace}
\usepackage{tikz}
\usetikzlibrary{shapes.geometric}
\usetikzlibrary{arrows, shapes}
\usepackage{seqsplit}

\newcommand{\code}[1]{{\ttfamily\seqsplit{#1}}}

\hypersetup{
	hidelinks = true 
}
\definecolor{blue(ncs)}{rgb}{0.0, 0.53, 0.74}
\newcommand{\MYhref}[3][blue(ncs)]{\href{#2}{\color{#1}{#3}}}%
\usepackage{float}

\definecolor{xtextBlue}{RGB}{42,8,255}
\newcommand{\BTCPP}{\code{BehaviorTree.CPP}\xspace}
\newcommand{\pytrees}{\code{py\_trees}\xspace}
\newcommand{\pytreesros}{\code{py\_trees\_ros}\xspace}
\newcommand{\skiros}{\code{SkiROS2}\xspace}

\usepackage{ifthen}
\usepackage{amssymb}
\newboolean{showcomments}
\setboolean{showcomments}{true} 
\ifthenelse{\boolean{showcomments}}
{\newcommand{\nb}[2]{
		\fcolorbox{gray}{yellow}{\bfseries\sffamily\scriptsize#1}
		{\sf\small$\blacktriangleright$\textit{#2}$\blacktriangleleft$}
	}
	
}
{\newcommand{\nb}[2]{}
	
}

\usepackage{fontawesome} 

\definecolor{findingcolor}{HTML}{2A7F8E} 

\newcounter{FindingIdx}
\newenvironment{finding}[1]{%
	\stepcounter{FindingIdx}%
	\par\medskip\noindent
	{\color{findingcolor}\textbf{Finding~\theFindingIdx\, --\, #1}}\par\nopagebreak
	\begin{tcolorbox}[
		blanker,
		borderline west={2pt}{0pt}{findingcolor},
		left=6pt, right=0pt, top=2pt, bottom=2pt,
		before skip=4pt
		]
	}{%
	\end{tcolorbox}
}

\begin{document}

\title{Behavior Trees for Robotic Systems: \\An Empirical Study on Practices and Experiences}

\author{Razan~Ghzouli, Jennifer~Horkoff, Daniel~Str\"uber and Rebekka~Wohlrab
}

\authorrunning{Ghzouli et al.}

\institute{
	R. Ghzouli  \at
	Department of Computer Science and Engineering,\\
	Chalmers University of Technology and University of Gothenburg,\\
	Gothenburg, Sweden\\
	\email{razan.ghzouli@gmail.com}
	\and
	J. Horkoff \at
	Department of Computer Science and Engineering,\\
	Chalmers University of Technology and University of Gothenburg,\\
	Gothenburg, Sweden\\
	\email{jennifer.horkoff@gu.se}
	\and
	D. Str\"uber \at
	Department of Computer Science and Engineering,\\
	Chalmers University of Technology and University of Gothenburg,
	Gothenburg, Sweden\\
	\textit{and} Department of Software Science,\\
	Radboud University, Nijmegen, The Netherlands\\
	\email{danstru@chalmers.se}
	\and
	R. Wohlrab \at
	Department of Computer Science and Engineering,\\
	Chalmers University of Technology and University of Gothenburg,
	Gothenburg, Sweden\\
	\textit{and} Software and Societal Systems Department,\\
	Carnegie Mellon University, Pittsburgh, PA, USA\\
	\email{wohlrab@chalmers.se}
}

\date{Received: date / Accepted: date}

\begin{center}
	\fbox{\parbox{0.9\textwidth}{\centering
			\textbf{Manuscript under review}\\[0.3em]
			\small This is a preprint currently under peer review. Content, results,
			and conclusions may change in subsequent versions.
			Please do not cite or distribute without the authors' permission.\\[0.3em]
			\textit{Version compiled: \today}
	}}
\end{center}
\vspace{1em}

\maketitle

\begin{abstract}
Over the last decade, behavior trees (BT) have become one of the dominating behavior models for coordinating missions of robotic systems in practice. 
Yet, empirical evidence regarding the adoption of BTs in real-world contexts remains limited, especially concerning practitioners' experiences and practices.
Without practitioner-grounded evidence from both academic and industrial settings, the research community risks developing guidelines and tools for BTs that are plausible in principle, but only partly aligned with the challenges encountered in practice. Furthermore, the scarcity of practitioner-grounded evidence from both academic and industrial contexts leads to ad-hoc practices, which impede software reuse, maintenance, and evolution.

To address this gap, in this paper, we report on the results of a mixed-methods study, combining a technical action research investigation at an automotive company with a survey  of 34 robotics practitioners.
Our results indicate that BTs improve team communication and the understandability of robotic decision-making logic, reflecting BTs practical value beyond mission coordination.
At the same time,  practitioners face multiple non-trivial design and integration decisions when adopting BTs in practice, which are complicated by the absence of adequate guidelines and tool support. 
The decisions span architectural and language mixing choices when implementing BTs and integrating them within ROS, for which we report followed patterns.
In addition, practitioners are faced with multi-factored granularity decisions, and mixed experiences with current libraries.
For the most used BT libraries, practitioners reported implementation challenges and documentation gaps.
As for planning algorithm usage for BTs, practitioners find  deciding the optimal BT node ordering challenging.
Practitioners views on scalability remained inconclusive, while they reported limitations in current libraries and practices for broader adoption.
We conclude our study with derived cross-cutting observations and provided implications for practitioners adopting BTs in robotic systems and researchers aiming to advance empirical understanding of BT adoption.


\end{abstract}


\section{Introduction}
Robots are already here! We see robots in warehouses, hospitals, and restaurants carrying out tasks such as logistics, disinfection, and food service, often autonomously around humans or in close collaboration with them.  In either case,  coordination between the various constituent skills is essential for reliable robotic behavior, and different approaches are available to implement this coordination in practice. Some approaches glue the coordination to the implementation of individual skills at a low-level of abstraction, i.e., coordination ``hard coded'' in the skills code, which often results in ad-hoc robotic systems that are difficult to maintain and reuse \citep{schlegel2010design}. Other practitioners adopt behavior modeling approaches which provide a higher-level of abstraction of skills coordination, thereby separating decision-making logic from low-level skills implementation \citep{dragule2025effects}. As robotics behaviors increase in complexity, the importance of the latter approach has grown \citep{filippone2026formalisms}. For decades, state machines were the dominant paradigm for behavior modeling in robotics. However, in the last decade, behavior trees (BTs) have become popular in robotics, and the two behavior models, state machines and BTs, are together the most widely adopted behavior modeling approaches in the robotics community \citep{street2024towards,%
	ghzouli2023behavior,%
	garcia2020robotics}.

Prior research has investigated BTs through controlled experiments, analyses of open-source projects, and individual researcher-led use cases \citep{colledanchise2016advantages, ghzouli2020behavior, hallen2024behavior}.  
Despite this growing body of work, our understanding of how practitioners adopt and experience BTs in concrete robotic development settings remains limited \citep{hallen2024behavior, colledanchise2021implementation}. In particular, existing evidence provides only partial insight into the software-engineering decisions that adopting BTs may entail. This includes how practitioners select and integrate BT libraries, connect BTs to Robot Operating System (ROS)-based systems, decide the granularity of the BT, reuse existing codebases, and communicate behavior logic across teams. 
These decisions are important because the benefits commonly associated with BTs, such as modularity, reactivity, readability, and reuse, do not arise solely from the modeling notation. Instead, these advantages depend on how BTs are implemented, integrated, and maintained in a given project context.
Without practitioner-grounded evidence from both academic and industrial settings, the research community risks developing guidelines and tools for BTs that are plausible in principle, but only partly aligned with the challenges encountered in practice. Furthermore, the scarcity of practitioner-grounded evidence from both academic and industrial contexts leads to ad-hoc practices, which impede software reuse, maintenance, and evolution \citep{garcia2023software}.

In this work, we provide an empirical study grounded in practice for building knowledge about BTs from both academic and industrial practitioners through a software engineering lens. We conduct an technical action-research study in an automotive company by introducing BTs and investigating the experiences faced during the process. We validate our observations during the technical action-research study and collect further data  by conducting a survey with 34 respondents targeting practitioners from both academia and industry. As a result, we provide real-world knowledge about BT adoption in practice exposing practices, experiences and influencing factors for different software engineering aspects.  To our knowledge, we offer the first study that evaluates BTs in practice using technical action-research, and the first study gathering a broad view of BT use and challenges in practices through a survey.

\subsubsection*{Goal and Research Questions}
We present a mixed method empirical study of BTs from academic and industrial practitioners perspectives. We formulated the following research questions:

\medskip
\noindent\textbf{RQ1.} \emph{What practices do practitioners follow when adopting behavior trees in robotic systems?}
We started by conducting a technical action-research study in an automotive company. Specific interest areas emerged during the technical action-research study that we wanted to investigate further with other practitioners from academia and industry. We conducted a follow-up survey targeting additional practitioners, asking about their practices concerning the interest areas noted during the technical action-research study. We collected quantitative data and reported the frequency for each aspect, along the data from the technical action-research participants. 
By reporting the different practices followed, we aim to build real-world knowledge about the usage of BTs from both academic and industrial practitioners that can inform future adoption decisions and development efforts.

\medskip
\noindent\textbf{RQ2.} 
\emph{What are practitioners experiences of adopting behavior trees in robotic systems?}
During the technical action-research study, we investigated the emerging  benefits and challenges when adopting BTs.
We introduced BTs to a team in an automotive company that was not familiar with them. For three months, one of the authors and the team worked on adopting BTs into a specific robotic system in which the team previously encountered challenges with their existing tool and its underlying behavior model
(shortly described in \cref{subsub:studycontext}). During and after the adoption of BTs, we identified interest areas, and reported the observed and experienced challenges and benefits. Finally, we used the survey targeting practitioners to validate the observations and collect further data. We collected qualitative data and used thematic analysis to code and analyze the data.
Our goal was to understand the emerging experiences when adopting BTs. By empirically identifying which benefits hold up in practice and which challenges were most widely experienced, we provided leads for potential improvements and set realistic expectations when adopting BTs.

\medskip
\noindent\textbf{RQ3.} \emph{What factors shape the design decisions of practitioners when adopting behavior trees in robotic systems?}
We identified system architecture and granularity modeling as essential design decisions during the technical action-research study. We surveyed the industry and academia practitioners about the influencing factors that shaped their decisions. We collected qualitative data and used thematic analysis to code the data. We reported the emerging codes, along the technical action-research participants factors. By understanding the driving factors behind the concerning design decisions, we wanted to establish an empirical foundation for developing design guidelines and enabling practitioners to make more informed choices.

\subsubsection*{Results}

Our study provides empirical knowledge grounded in practice that has previously been lacking in discussions of BTs. 
We report followed practices, experiences influencing factors 
to support practitioners in making informed adoption decisions and guide future improvements efforts. We report findings across eight interest areas spanning BT design, integration, and use in practice: granularity of BT model, BT libraries, BT-to-ROS nodes architectural design, mixing programming languages during the programming of BTs, team communication, understandability, the usage of planning algorithms with BTs and the scalability of BTs. Finally, we highlight observations emerging across the interest areas to contextualize the individual findings, and we derive implications for practitioners adopting BTs in robotic systems and researchers studying them empirically.
In the following, we highlight briefly the results (see \Cref{sec:results} and \Cref{sec:discussion} for details).

Regarding BT model granularity, no single level emerged as the majority preference, and practitioners recommended a trial-and-error mindset due to the complexity of the decision. Most respondents used \BTCPP and \pytrees, describing them as straightforward and easy to use, respectively, while reporting implementation challenges faced using them. The transition from BT modeling to implementation within ROS was challenging. In the absence of established guidelines, practitioners had to decide wither to implement BT using one programming language or mixing languages, and how to architect BT within ROS nodes. We observed patterns regarding the former practices and reported them. The majority of practitioners did not use a planning algorithm to automatically generate or modify BTs.  Practitioners reported challenges in optimizing node ordering without a planning algorithm, yet there was no consensus on its necessity.

Beyond implementation, practitioners reported that BTs facilitated better communication among team members and non-technical stakeholders and enhanced the clarity of robotic decision-making logic. Finally, practitioners experiences regarding BT scalability were mixed and inconclusive, highlighting limitations in current libraries and practices that were viewed as insufficient for broader adoption.

\subsubsection*{Perspectives}

With this work, we hope to bring knowledge and starting points for improvements to two communities, software engineering and robotics. For software engineers in robotics, practitioners perspective on libraries and design decisions, including architectural patterns and granularity, may inform the development of guidelines for best software practices to address the reported challenges. For roboticists, the highlighted experiences regarding planning algorithms and ROS integration, may inform the development of improved integration frameworks and algorithms. 
 
We provide an online appendix containing the survey instrument, and anonymized dataset as a replication package to facilitate further research \citep{appendix:online}. The dataset contains detailed statements by practitioners describing the reported challenges, benefits and influencing factors, in addition to the highlighted thematic analysis coding.


\section{Background}

In the following, we provide brief introduction about behavior trees (BTs) and the Robot Operating System (ROS) to familiarize our reader with concepts discussed later in the paper. 

\subsection{Behavior Trees}
With the widespread adoption of robotics across domains, robotic missions have grown in complexity and frequently require the execution and coordination of multiple tasks.
These tasks are further decomposed into actions, also referred to as skills, necessitating precise coordination to achieve reliable mission outcomes.
To manage this coordination, various control structures have been used to model and coordinate robotic behavior, including behavior trees, state machines, Petri nets, subsumption architecture, and hierarchical task networks, each with its benefits and limitations \citep{colledanchise2018behavior}. 

In the last decade, BTs became one of the most used behavior model in robotics projects \citep{ghzouli2023behavior}. In addition, the Robot Operating System (ROS), a leading framework in robotics, has incorporated BTs within its second version ecosystem (ROS 2) to facilitate mission coordination \citep{macenski2020marathon}. Similarly, PlanSys2, the ROS 2 planning system, utilizes BTs to execute plans generated by a Planning Domain Definition Language (PDDL)-based task planner \citep{martin2021plansys2}. 

\begin{figure}
	\begin{center}
		\includegraphics[
		width=\linewidth,
		trim=30mm 90mm 60mm 60mm,
		clip
		]{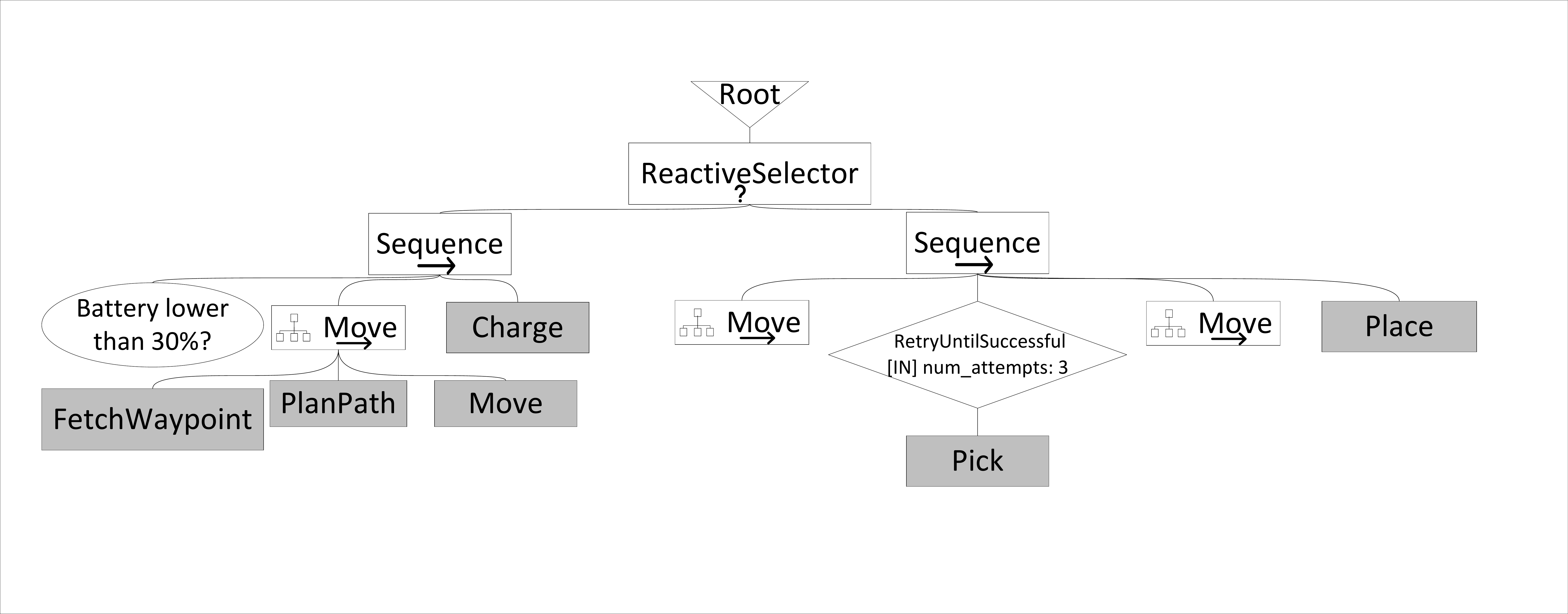}
	\end{center}
	\caption[BT example]{\footnotesize{An example of a pick \& place mission represented in BTs.}}%
	\label{fig:BTExample}
	
		\vspace{-3mm}
	
\end{figure}

\noindent\textbf{BT example:} To showcase BTs, \Cref{fig:BTExample} presents an example of a pick and place mission modeled in BTs with a battery-aware fallback behavior.
The top-level behavior uses a \code{ReactiveSelector} that always checks battery status before running other tasks. 
The left branch is a recharging \code{Sequence}, guarded by a condition nodes \code{Battery lower than 30\%?}.
If the battery level is below 30\%, the robot obtains the waypoint of the charging station \code{FetchWayPoint}, plans a path to it \code{PlanPath}, navigates to the charging station \code{Move}, and executes the Charge action until the battery is full \code{Charge}. Otherwise, the robot performs the pick-and-place \code{Sequence} branch: navigates to the pick station through the move sub-tree \code{Move}, retries up to three times using \code{RetryUntilSuccessful} to pick an object \code{Pick}, then moves to the place station \code{Move} to place the object \code{Pick}. If the battery drops mid-task, then the robot moves to the charging station and the robot resumes pick-and-place after recharging, skipping completed steps. The \code{Move} sub-tree is reused for navigation to any target by passing the location as a parameter (charging, pick or place station).

\noindent\textbf{BT concepts:} A BT is a directed tree with a root node as the entry point. BTs have two types of nodes non-leaf nodes, called control-flow nodes, and leaf nodes, called execution nodes. Control-flow nodes have four main types: Sequence, Selector, Decorator and Parallel. The execution nodes have two types: action nodes and condition nodes.
The action nodes encapsulate robotic skills (e.g., \code{Pick}), while conditions test conditions (e.g., \code{Battery lower than 30\%?}). The semantic of control-flow nodes coordinates the execution of all tree nodes.
Execution proceeds by sending a signal called tick from the root downward at a fixed frequency, and each ticked node returns one of three statuses to its parent: success, failure, or running \citep{colledanchise2018behavior}. Through the recursive propagation of ticks and statuses, the control-flow nodes compose the execution nodes into the complete mission behavior.

In the pick-and-place example, the main control-flow nodes are Sequence, a type of selector \code{ReactiveSelector} and a type of decorator \code{RetryUntilSuccessful}. 
A sequence node needs all its children to succeed for it to success, whereas a selector node succeed as soon as any one of its children succeeds. Parallel nodes execute all their children concurrently. However, the robotics BT libraries do not implement concurrent execution. They implement parallel nodes as a generalization for sequence and selector nodes by exposing a configurable success policy \citep{ghzouli2020behavior}.
Decorator nodes enable richer control flow such as retry and for loops, enabling developers to define custom decorators tailored to their application. For example, the decorator \code{RetryUntilSuccessful} lets the robot try picking up to three times if it fails. 
Current robotics BT libraries offer special types of decorator, sequence, selector and parallel nodes. Notably, the semantics of control-flow nodes are not standardized across libraries \citep{bernagozzi2025model, ghiorzi2024execution}.  

The two most used BT libraries are \BTCPP and \pytrees \citep{ghzouli2023behavior, iovino2022survey}. \BTCPP\footnote{\MYhref{https://www.behaviortree.dev/}{behaviortree.dev}} is a C++ library that has a graphical user interface (GUI) called Groot. Groot allows modeling and modifying BTs, and monitoring a running BT. Groot runs independently without the need to setup a robotic system or connecting it to ROS, allowing users to start modeling faster in BTs. \pytrees\footnote{\MYhref{https://py-trees.readthedocs.io/en/devel/}{py-trees.readthedocs.io}} is a Python library, which is extended to \pytreesros\footnote{\MYhref{https://py-trees-ros.readthedocs.io/en/devel/}{py-trees-ros.readthedocs.io}} to support ROS. A viewer is available, called PyTrees ROS Viewer\footnote{\MYhref{https://github.com/splintered-reality/py_trees_ros_viewer}{github.com/splintered-reality/py\_trees\_ros\_viewer}}, to only visualize BTs. However, it is not well maintained\footnote{By April 2026, last repo update was Jan 2025.}. 

\subsection{The Robot Operating System}

The Robot Operating System (ROS) is a widely used open-source middleware and frame for developing robotics applications, and it offers a set of software libraries for communication, software abstraction and others for building distributed robotics applications \citep{quigley2009ros}. ROS facilitates modular robotic system design by organizing applications into nodes that communicate by exchanging typed messages through well-defined interfaces, such as topics, services, and actions. A ROS node is usually responsible for a specific functionality and communicating with other nodes through ROS interfaces. A node could be spread across one process or distributed across multiple processes.

ROS supports several communication mechanisms, including publish/subscribe with topics, synchronous request/reply via services, and goal-based actions for long-running tasks. In goal-based actions, ROS provides actions, a client sends a goal to a server, which performs the task, optionally provides feedback, and returns a result upon completion. This suits robot tasks like navigation or manipulation that benefit from progress updates. 

The reader of our paper needs to distinguish between a BT node, ROS node, a tree action and a ROS action.
Throughout this paper, the term BT node refers to any node within a BT (control-flow nodes or execution nodes), while a ROS node refers to the modular computational unit within a ROS system, as described earlier.  
By tree action, we refer to the execution-node type in BTs, whereas by ROS action we refer to the ROS communication interface between an action client and an action server.

Currently, there are two versions of ROS, ROS 1 and ROS 2. While core ideas are similar, ROS 2 differs from ROS 1 mainly in its underlying communication architecture, offering improved support for real-time, secure, and scalable robotic applications \citep{diluoffo2018robot}.
ROS 2 offers official client libraries for C++ (rclcpp), Python (rclpy) and RUST (rclrs), enabling seamless cross-language communication. In this paper, ROS 2 was used and the availability of the  official client libraries supported mixing programming languages within BTs (discussed further in \cref{subsec:BTMixingLang} and \cref{subsec:ROSBTArch}). 

ROS is a community built and driven ecosystem, and the software robotics community actively contribute to it \citep{kolak2020takes}. Practitioners exchange knowledge through dedicated website and forums such as the ROS Wiki and ROS Discourse \citep{estefo2019robot}. ROS wiki\footnote{\MYhref{https://wiki.ros.org/}{wiki.ros.org}} is the main website for ROS to publish libraries, packages, tutorials and for practitioners to look for community support. ROS Discourse\footnote{\MYhref{http://discourse.ros.org/}{discourse.ros.org}} is a forum site and communication channel widely used by practitioners in the robotic community for announcing updates and discussing topics. In this paper, we used ROS wiki to find tutorials about BT-ROS implementations, and we used ROS Discourse to reach a wide range of practitioners for the survey.


\begin{figure}
	\begin{center}
		\includegraphics[
		width=\linewidth,
		trim=198mm 100mm 105mm 10mm,
		clip
		]{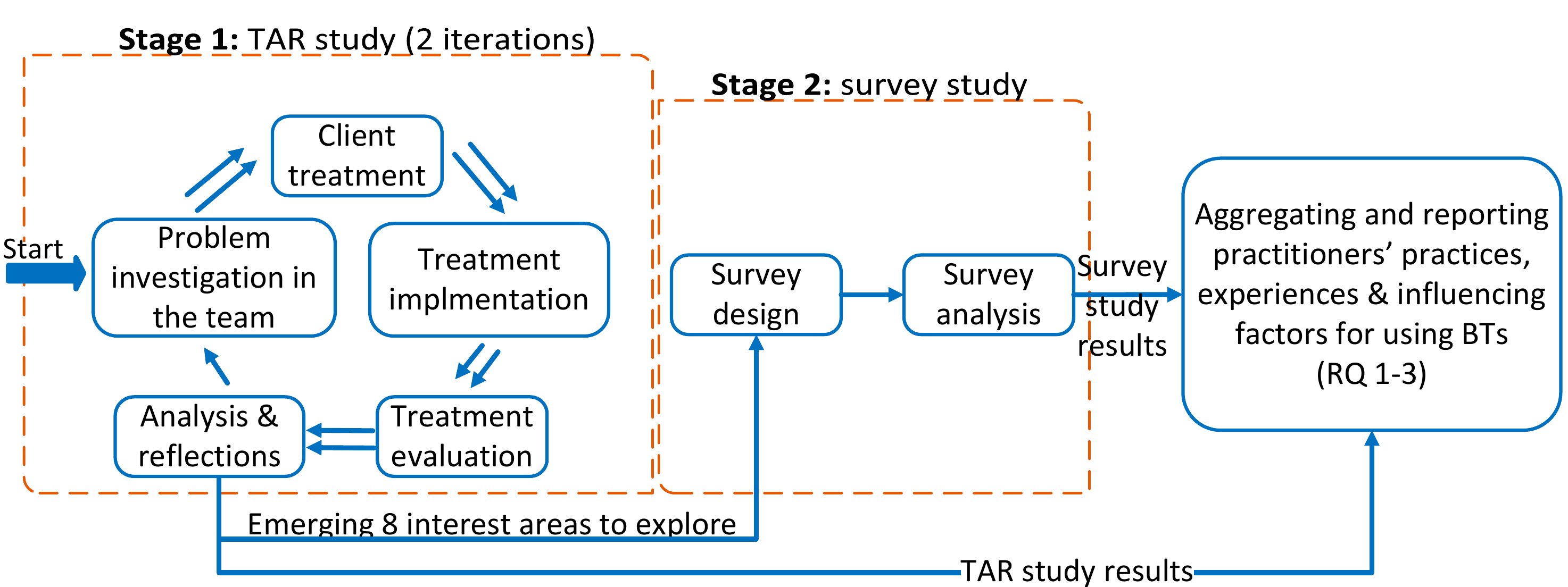}
	\end{center}
	\caption[Research method]{\footnotesize{An overview of the research methodology.}}%
	\label{fig:method}
	
	\vspace{-3mm}
	
\end{figure}

\section{Research Method}
\label{sec:methodology}

We followed a similar design to the concurrent triangulation design mentioned by  \citet{creswell2003advanced}. 
We used two-stage mixed methods to understand the problem and collect our data. \Cref{fig:method} shows an overview of our followed research method. The first stage was a technical action-research (TAR) study conducted at a company to explore and understand the experiences of practitioners. We collected and analyzed qualitative data, which resulted in preliminary observations and eight emergent interest areas that could not be fully characterized within a single organizational context.
We wanted to expand the results from a single organization context with a second stage  survey study with academic and industry practitioners. We used the emerging interest areas to design the survey questions, and collected qualitative and quantitative data to support or refute the observations during the technical action-research study. Furthermore, we collected additional information about the interest areas to enrich our results with experiences and practices.
The two phases are complementary. The technical action-research results provided depth and contextual grounding, whereas the survey results extended the scope of inquiry and added breadth. 
By conducting the mixed methods, we gained perspectives from the different types of data, confirming initial observations and generating new findings.
In the following, we explain further the two stages.

\subsection{Stage 1: Technical Action-Research Study}
We conducted a technical action-research in our investigation of BTs adoption in industrial context following a similar structure to \citet{wieringa2014design} technical action-research. We adapted Wieringa's technical action-research structure  to our study, which is an acceptable practice in empirical software research \citep{kampenes2008flexibility, sjoberg2007future, molleri2024teaching}. The technical action-research design is reflected in the method choices presented in this section.

We chose technical action-research since it is an artifact-driven approach and supports investigating an artifact within a real-world context by actively introducing the artifact into practice to solve a relevant problem \citep{wieringa2012technical}. We chose technical action-research over action-research method since action research is a problem driven, while our study is an artifact-driven problem. We chose technical action-research over a use-case study because technical action-research involves active researcher participation, while use-case studies examine an artifact in a real-world context but researchers remain passive and only observe how the artifact is adopted \citep{runeson2009guidelines}. Moreover, researchers in a technical action-research study are actively involved in applying the artifact in the real-world context while simultaneously investigating to build knowledge \citep{wieringa2014design, wieringa2012technical}, fitting our study goal. 
Based on \citet{wieringa2014design}, technical action-research studies span over three-level of structure which are a design cycle, an empirical cycle and a client engineering cycle.

In our study, BT is the artifact, well established, but has limited empirical evidence about its adoption in industrial settings and practitioners experiences. The treatment was adopting BTs in an industrial context at the company's team. We were actively involved in the adoption of BTs into the industrial context within the company's team. We conducted a client engineering cycle and an empirical cycle, noting that the client engineering cycle is usually intertwined with the empirical cycle \citep{wieringa2014design}. We did not conduct a design cycle since the investigated artifact in our case is well established and it was out of our scope to redesign it \citep{wieringa2014design}. 
We conducted two iterations of the client engineering cycle. 
By introducing BTs into the team, we supported company A to adopt a new behavior model approach and build experience. For our study, we investigated and built knowledge from the results about practices, experiences, and factors that influence design decisions. 

In the following, we provide details about the technical action-research study in terms of the study context, how we collected and analyzed data, and the different iterations. Except for \Cref{sec:methodology}, we use the term action research throughout the remainder of the paper instead of referring to the specific type for brevity. 

\subsubsection{Study Context}
\label{subsub:studycontext}
We provide information about the industrial context in which our research was conducted in it for better understanding of the results \citep{petersen2009context, wieringa2014design}.
In the following, we provide context about the team in the company that we collaborated with. In addition, we provide information about the current practices in the team for mission specification and their perception about BTs before starting the study. We collected the latter information during a first visit to the team.

\noindent\textbf{The team:} We performed the technical action-research study with a team at an automotive manufacturer (Company A) in Sweden that contacted us.
Two people were actively involved throughout the cycles and one additional person was only involved in the first diagnosis phase.
The first person is a safety expert and researcher at Company A (P1). The second person is an industrial researcher actively involved in Company A projects (P2), but works in a different organization (Company B). 
The third person is a robotics's expert and researcher at Company A who was involved only in the diagnosis phase of the first cycle (P3). None of the involved people from had previous knowledge of BTs.

\noindent\textbf{Visiting arrangement: } Two of the paper authors started the study with a visit to the team from company A, including the safety expert (P1) and robotics expert (P3). We treated the first visit as a problem investigation and conducted a workshop \citep{safsten2020research} with P1 and P3. 
After discussion, the first author visited the company once a week for three months (Sep-Nov) and collaborated with P1 and P2 to adopt BTs and observe emerging experiences. 
Through out the paper, we call that first author the visiting researcher.

\noindent\textbf{Participants practices for mission specifications \& their challenges:} 
We had a first visit to the team who contacted us to understand their current practices for mission specifications and the faced challenges (client problem investigation). 
For mission specifications, the involved team at Company A used an open-source and in-house developed tool that combines a planning algorithm with state machines called Sequence Planner (SP) \citep{dahl2022sequence}. P3 was was involved in its development. SP uses state machines for behavior modeling and control coordination of the robotics missions. The SP framework uses a planning algorithm for online planning and generating the control logic in state machines. The involved developers wanted flexible mission specifications with no hard-coded sequences. Thus, SP requires defining pre-conditions and post-conditions for actions and giving the planner goals. Depending on the active goal, the planner generates an appropriate sequence of actions. 

P1 was not involved in the development team of the SP tool.
She reported that those who were not involved in the development of the tool found it hard to start working with it and program due to lack of documentation. Also, P1 reported that it was not intuitive to understand the robotics missions on a high level or to follow the decision logic behind the planner's generated sequences, which made risk assessments and discussions across teams difficult. Handling fallback behavior was also limited. Although SP's planner could handle fallback by re-planning, fallback behavior was not explicitly represented in the behavior model (state machine). Finally, there used to be a GUI for the tool. However, it was hard to maintain the GUI as the tool was evolving, resulting in obsolescence GUI and no available visualization of the robotics missions. 

With the advancement of AI in the company and collaborative robotic missions with operators, P1 highlighted that mission visibility was essential. With multiple unpredictable scenarios and various people involved, clear visibility of mission progress and decision logic was critical. 
Although SP used a planner, having stat machine as the underlying behavior model limited reactivity. Reactivity was tied to action failure rather than to continuous environments monitoring, which limited responsiveness in missions involving operators and dynamic environments.
In addition, robotics projects typically involved multiple stakeholders such as engineers, safety experts, technicians, maintainers, and operators. With so many contributors, maintaining mission visibility ensures everyone understands the project’s status and requirements. Thus, in such diverse and complex projects, clear communication between stakeholders and understanding of the mission components were essential for successful projects.

\noindent\textbf{Participants perception of BTs:}
In the first visit to the team (first diagnosis phase), we discussed the opinions of P1 and P3 about BTs after an introductory presentation. P1 had no knowledge about BTs, while P3 knew about the existence of BTs as a behavior model but never used them. Both P1 and P3 thought BT model provided an overview of a robotics mission and that the availability of domain-specific languages (DSLs) to implement them with a graphical-user interface (GUI) was positive.  

In general, P1 and P3 had different initial perceptions about BTs. P3 saw BTs as a hard-coded model for a mission sequence and when mission flexibility was not needed then BTs might be useful. He perceived BTs as good fit for toy examples and that BTs do not scale well. Finally, he expressed that they have been thinking of using BTs when flexibility in the missions' planning was not a priority, but was discouraged with the assumption that the programming of BTs was not easy.  
P1 saw potential in using BTs for visualizing missions and communicating with involved stakeholders, especially robotic operators to understand active parts of a mission. However, she saw potential limitation in the visualization depending on the mission size.

\subsubsection{Data Collection and Analysis}
We collected data from the problem investigation, client treatment's design, implementation and evaluation, and research execution in the cycles. We used workshops and notes in a diary forms to collect data \citep{safsten2020research, wieringa2014design}. 
During the recurrent visits to the team, we collected data through observations and taking field notes in a diary, which is a common data collection form in technical action-research studies \citep{wieringa2014design}. Furthermore, the visiting researcher conducted workshops with P1 and P2 following each iteration (evaluation step), serving both as a forum for discussion and evaluation.
In the workshops, we presented our observations and asked follow up questions to participants, and during the discussions the visiting researcher collected highlights and notes. Our goal in using this data collection form was to address specific questions, while allowing participants to freely explore other emerging observations \citep{safsten2020research}.

The visiting researcher analyzed the collected data from the field notes' diary using inductive content analysis \citep{vears2022inductive}. She read through diary notes to get familiarized with the data, extracted and organized key statements from the diary and workshops' presentations in digital mind maps using Miro\footnote{\MYhref{http://miro.com/}{miro.com}}. The analysis process was iterative through out the technical action-research iterations to identify emerging patterns within the data.
Eight interest areas emerged from the analysis and key observations about them. 
We conducted discussions with three of the authors, including the visiting researcher, focused on enhancing and clarifying the emerging interest areas rather than validating.
Our followed data analysis process supported the exploratory and reflexive nature of our technical action-research study \citep{vears2022inductive}. We followed the process to allow patterns and observations to emerge directly from the data without imposing a predetermined patterns.
We present the interest areas and the results in \Cref{sec:results}.

\subsubsection{Iterations of Client Engineering Cycle}
We conducted five steps spanning the empirical and client engineering cycles: research execution, problem investigation, client treatment's design, implementation and evaluation. 
We do not elaborate on the research execution as separate technical action-research components since it corresponds to the client engineering cycles described below \citep{wieringa2014design}.
We conducted two iterations of the client engineering cycle within the team. We followed similar structure to \citet{wieringa2014design} and adapted the method to our study, which is an acceptable measure in empirical software research  \citep{kampenes2008flexibility, sjoberg2007future, molleri2024teaching}. To describe each iteration, we use similar terms to  \citet{wieringa2014design} with adaption to our study. The first iteration involved understanding the team context and problem, as well as introducing BTs to the team context. 
We noticed a need to adapt the treatment design based on the first iteration evaluation, thus we did a second iteration and refined assumptions regarding BTs adoption.
In the following, we present details about the two iterations.

\subsubsection*{Iteration 1}
\noindent\textbf{Problem investigation in the team:}
In the first visit to the company, one of the researchers lead a workshop, while the other researcher took notes. P1 and P3 were present from the company side. We presented BTs in terms of basic concepts, the reason for using behavior models such as BTs and our goal for understanding the industry perspectives about adopting BTs.
A discussion was followed to understand the current practices at the company to model robotics missions, challenges and the participants' perception about BTs.
After discussions with P1 and P3, we decided collectively to have the visiting researcher with P1 and P2 investigate the adoption of BTs into a robotic system.

\noindent\textbf{Treatment design and implementation:} Following data analysis of the problem investigation step and discussions with P1 and P2, the first iteration treatment focused on introducing P1 and P2 to BT model, relevant programming libraries and implementing BTs in their robotic system. We spent two of the visits implementing the treatment.
To implement the treatment, the visiting researcher conducted an introductory presentation about BT models by introducing basics concepts, followed by an example of a robotic mission which was then modeled by P1 and P2 with BTs. The visiting researcher presented existing libraries for implementing BTs, highlighting their support for ROS, availability of a GUI, maintenance status, and supported programming languages, such as Python and C++. The presentation focused on \pytrees and \BTCPP since they are the two most used libraries for BTs and are actively maintained. The presentation was followed by a discussion and questions about BT model and the libraries. P1 and P2 were interested in "pick and place" missions since the robotic system involved a robotic arm with similar tasks. Thus, the visiting researcher provided two different open-source projects using \BTCPP and \pytrees, respectively, for pick and place missions. The projects were picked from an available collection of open-source projects \citep{ghzouli2023behavior}. The projects were given to P1 and P2 to familiarize themselves with the libraries. 
P1 and P2 decided to use \BTCPP version 4.6 since it has a GUI, support ROS and is well maintained.

\noindent\textbf{Implementation evaluating:} The visiting researcher called for a treatment evaluation  after analyzing her observations from taking notes in the diary.
The visiting researcher noticed that integrating BTs from scratch was more complicated than expected. The initial scope was broad and did not fully consider practical challenges such as reusing existing code and working and integrating a new BT library into an established software and hardware robotic system. She also noticed that the treatment goals did not account for real industrial settings.

The visiting researcher conducted a workshop with P1 and P2 to evaluate the treatment and collect their perceptions. During the workshop, she presented an overview of the technical action-research cycles. She presented the problem in hand and the followed treatment. In addition, she presented her observations to date and asked P1 and P2 whether they agreed with the observations and their experiences.
After discussion, the visiting researcher, P1 and P2 decided to refine the treatment and continue further into a next iteration. 
We report the learning from iteration one in \Cref{sec:results} cumulatively with iteration two learning. 

\subsubsection*{Iteration 2.}
\noindent\textbf{Treatment redesign and implementation:}

Following reflection on participants' observed and reported experiences in iteration one, P1, P2 and the visiting researcher collectively decided to narrow the scope and prioritized integrating the selected BT library into the existing system rather than building from scratch. The first iteration revealed that the initial treatment did not fully account for the practical needs of industrial BT adoption, including the need to reuse existing code and integrate with established software and hardware infrastructure.

During the second iteration, P1, P2, and the visiting researcher collaborated to integrate the BT library and use it to model and execute a pick-and-place mission for controlling the robotic system known as Robot in the Air (RITA) \citep{erHos4157292framework}. 
RITA is an intelligent automation system usually used for missions of assembling kits in assembly lines. The robotic missions for RITA at company A usually involved a robotic arm on a gantry that pick items from boxes and place them on a mobile robot that deliver items to an operator, and vise versa.
Participants used to use the SP tool, with state machines as the behavior model. With the advancement of AI at company A and evolving collaborative robotic missions with operators, certain aspects grew in importance. 
P1 expressed the importance of clarity of decision logic and visualization of it, reactive fallback behavior and improve communication across teams. The SP tool fell short on these aspects.

P1, P2 and the visiting researcher investigated the introduction of BTs for controlling RITA, motivated by the former aspects, to assess associated benefits and challenges. The former aspects informed the investigation, but did not fully define the scope of it. The goal was to focus on the integration of chosen BT library into the existing robotic system and hardware resources. 
The pick-and-place mission involved moving the robotic arm to three different locations using a tree action node \code{Move} with the location as a variable parameter. In addition, a pick and place tree nodes were involved in the mission, \code{PickObject} and \code{PlaceObject}. 
The plan was to use Ubuntu version 22.04 LTS (Jammy Jellyfish) and ROS version 2 (IronROS) to connect with existing hardware resources and their drivers.

In the treatment implementation, P1, P2 and visiting researcher carried out the plan. The majority of planned visits were during this step. P2 implemented the majority of the plan with the support of P1, and the visiting researcher supported with technical knowledge.
 
\noindent\textbf{Implementation evaluation:} The implementation evaluation of iteration two was done in similar way as iteration one. The visiting researcher analyzed her observations and notes and conducted a workshop with P1 and P2 to evaluate and collect data. 
We report the results in connection to iteration one and the survey results in \Cref{sec:results}.

\subsection{Stage 2: Survey Study}
We conducted the survey study to enrich and cross-validates our observations from the technical action-research study. Survey studies offer breadth than a single industrial study can provide \citep{safsten2020research}, which was our goal from conducting it.
We selected a survey instead of other data collection methods, such as interviews, to reach a large and diverse group of practitioners from both academia and industry within a manageable time. 

We wanted to enrich and cross-validate our findings from the technical action-research study by investigating whether other practitioners encounter similar practices, challenges, and design considerations when adopting BTs in practice.
In the following, we report the design of the survey, how we targeted and reached out to respondents, and survey data analysis. Finally, we provide an overview of the respondents information after analyzing the data.
The full survey and anonymized data are provided in our online appendix \citep{appendix:online}.

\subsubsection{Targeted Respondents:} Finding a sample that would generalize to the population of practitioners using BTs is challenging. We do not  know accurately the number of practitioners using BTs in the world, nor how to reach them. We used social platforms to maximize our reach to practitioners \citep{storey2014r,begel2010social}. We used the social platform LinkedIn to reach existing connections and the ROS Discourse for wider reach to robotic practitioners. We do not claim that the used social platform provided access to all BTs practitioners, but following our knowledge of robotics and BTs practitioners' platforms, they provided adequate access to the community.

\begin{table}[!tbp]
	\caption{Survey questions on participant background.}
	\label{tab:survey-questions-participant}
	\centering
	\scriptsize
	\setlength{\tabcolsep}{4pt}
	\renewcommand{\arraystretch}{1.0}
	\begin{tabularx}{\textwidth}{@{}p{1.5cm} X p{2.2cm}@{}}
		\toprule
		\textbf{Q\#} & \textbf{Survey question} & \textbf{Data type} \\
		\midrule
		Q1  & Familiarity with BT key concepts                  & Nominal \\
		Q2  & Industry or academia affiliation                  & Nominal \\
		Q3  & Programming languages used in robotics            & Open-ended \\
		Q4  & Years of experience in robotic projects           & Ordinal \\
		Q5  & Prior use of a BT model                           & Nominal \\
		Q6  & Context(s) of BT usage                            & CATA \\
		Q46 & Participant name (contact information, optional)  & Open-ended \\
		Q47 & Participant email (contact information, optional) & Open-ended \\
		\bottomrule
	\end{tabularx}
\end{table}

\begin{table}[!tbp]
	\caption{An overview of the RQ-mapped survey questions.}
	\label{tab:survey-questions-rq}
	\centering
	\scriptsize
	\setlength{\tabcolsep}{3pt}
	\renewcommand{\arraystretch}{0.95}
	\begin{tabularx}{\textwidth}{@{}p{1cm} p{1.5cm} X p{1.8cm}@{}}
		\toprule
		\textbf{RQ} & \textbf{Q\#} & \textbf{Survey question} & \textbf{Data type} \\
		\midrule
		\multicolumn{4}{@{}l}{\textit{Granularity of BT model}} \\
		RQ1 & Q30 & Granularity level typically used          & Likert (0--5) \\
		RQ2 & Q29 & Struggle with BT granularity decisions    & Nominal \\
		RQ3 & Q31 & Factors affecting granularity choice      & Open-ended \\
		\addlinespace[2pt]
		\midrule
		\multicolumn{4}{@{}l}{\textit{BT libraries}} \\
		RQ1 & Q7  & Library used for creating BTs             & CATA \\
		RQ1 & Q8  & Library with the most experience          & Nominal \\
		RQ2 & Q9  & Reason for not using any tool/library     & Open-ended \\
		RQ2 & Q10, Q13, Q16, Q19, Q23 & Most experience library: setup documentation ease   & Likert (0--5) \\
		RQ2 & Q11, Q14, Q17, Q21, Q24 & Most experience library: documentation completeness & Likert (0--5) \\
		RQ2 & Q12, Q15, Q18, Q22, Q25 & Most experience library: overall experience         & Open-ended \\
		\addlinespace[2pt]
		\midrule
		\multicolumn{4}{@{}l}{{\textit{Language mixing within BTs \& BT-to-ROS nodes architecture \& Planning algorithms}}} \\
		RQ1 & Q28 & Language mixing vs.\ rewriting approach   & CATA \\
		RQ1 & Q26 & BT-to-ROS architectural pattern(s) encountered & CATA \\
		RQ3 & Q27 & Factors influencing BT-to-ROS pattern choice        & Open-ended \\
		RQ1 & Q32 & Use of a planning algorithm with BTs      & Nominal \\
		RQ1 & Q33 & Type of planner used                      & Open-ended \\
		\addlinespace[2pt]
		\midrule
		\multicolumn{4}{@{}l}{\textit{Communication \& understandability \& scalability}} \\
		RQ2 & Q34 & Effect of BTs on stakeholder communication & Nominal \\
		RQ2 & Q35 & Elaboration on communication effect       & Open-ended \\
		RQ2 & Q36 & Effect of BTs on mission understandability & Nominal \\
		RQ2 & Q37 & Elaboration on understandability effect   & Open-ended \\
		RQ2 & Q39 & Elaboration on scalability experience     & Open-ended \\
		\addlinespace[2pt]
		\midrule
		\multicolumn{4}{@{}l}{\textit{Evaluating predefined statements}} \\
		RQ2 & Q38 & BTs scale effectively with increased system complexity & Likert (0--5) \\
		RQ2 & Q40 & Transitioning from BT model to ROS implementation is challenging & Likert (0--5) \\
		RQ2 & Q41 & BT library and BT--ROS integration documentation are lacking & Likert (0--5) \\
		RQ2 & Q42 & Challenging to choose BT-to-ROS architectural design  & Likert (0--5) \\
		RQ2 & Q43 & Challenging to optimize execution order without a planner & Likert (0--5) \\
		RQ2 & Q44 & High-granularity BT limits flexibility and maintainability & Likert (0--5) \\
		RQ2 & Q45 & BTs improve readability by structuring decision-making logic & Likert (0--5) \\
		\bottomrule
	\end{tabularx}
\end{table}

\subsubsection{Survey Design:} 
Our goals with the survey were to evaluate observations from the technical action-research study, and to report insights about respondents' experiences, practices and influencing factors. Eight interest areas emerged during the technical action-research study that influenced our survey design, including: granularity of BT models, BT libraries, BT-to-ROS nodes architectural design, mixing programming languages during the programming of BTs, team communication, understandability, the usage of planning algorithms with BTs and the scalability of BTs.     

\Cref{tab:survey-questions-participant} and \Cref{tab:survey-questions-rq} shows an overview of the survey questions, collected data types and their relations to our research questions.
Overall, the survey had three parts. The first part collected respondent information about the domain of work, years of experience with robotic projects and BTs. We added two optional questions to collect respondents contact information in case of a follow-up study. The second part focused on experiences, practices and influencing factors when adopting BTs in projects, covering the eight interest areas. 
Finally, we asked respondents to report their agreement levels on predefined statements concerning challenges and benefits from the technical action-research observations. 

We designed the survey to include likert‑scale questions for agreement levels, open-ended questions and check-all-that-apply (CATA) questions for usage patterns. To illuminate answers from respondents with no prior experience with BTs and capture actual experiences, not  speculation, we implemented two questions in different stages of the survey to discard  answers from those who had not used BTs. We also had "I don't know" and "Not applicable" choices for all questions. 

We informed participants of the study's purpose, data retention period, and protections for anonymity prior to their responses. Participation was voluntary, and completing the survey implied consent. Participants could withdraw at any time by closing the survey window.

We piloted the survey with three respondents from three different organizations and companies. The pilot allowed us to refine two of the questions to reduce ambiguity and confirm the estimated duration for taking the survey (15-20 minutes). After refinements, we reached out to the pilot respondents to confirm the effectiveness of the changes. Due to the scarcity of respondents, we confirmed with the pilot group that the changes did not affect their answers, then requested their consent to use their results in the final study. Finally, we made the survey publicly open for two months.

\subsubsection{Data Analysis:} 

We had 39 practitioners responding to our survey. We discarded five of them using our filtering questions for capturing those with no-prior experience with BTs. The data analysis was based on the answers of 34 practitioners. Not all open-ended questions (elaborative questions) received responses, thus, we highlighted in the results the number of responses received (n). We also highlighted in the figures, when possible, the number of received responses and the number of responses for "not applicable" (NA) and "I don't know".

The survey collected qualitative and quantitative data.
The quantitative data were collected from the nominal, CATA and likert‑scale questions.
We used descriptive statistics to present the quantitative data following best practices for studies with exploratory nature to understand practices and real-world settings \citep{stol2018abc}. The qualitative data were collected from the open-ended questions.
We conducted inductive-thematic analysis on the qualitative data regarding influencing factors and experiences of respondents for each relevant question \citep{terry2017thematic}. One of the researcher coded responses, followed by conducting a workshop with another researcher that was not involved in the original coding to verify codes. When a response addressed multiple themes, we segmented the response and coded each relevant segment separately under its corresponding thematic code.
For each question, we discarded responses that did not answer the question. The thematic codes along the data and discarded responses for each question are highlighted in our online appendix.

\subsubsection{Respondents Information}

\begin{figure}[t]
	\begin{center}
		\includegraphics[
		width=\linewidth
		]{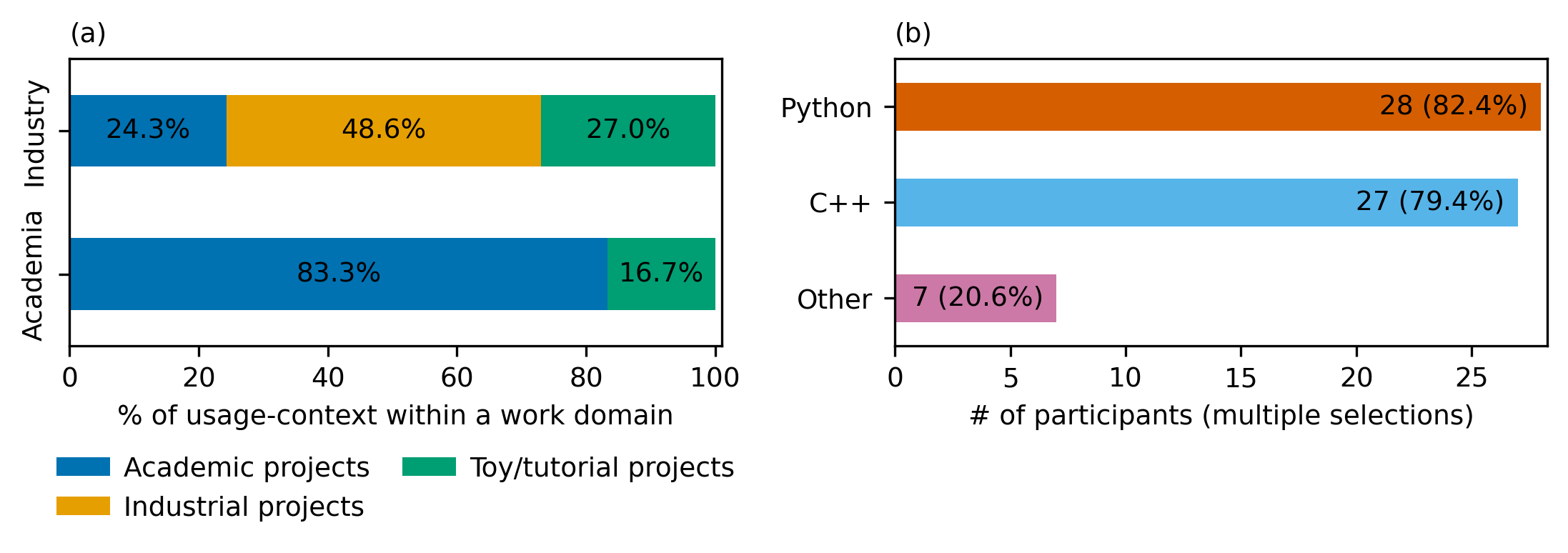}
	\end{center}
	\caption[Respondents info]{\footnotesize{Survey respondents results about (a) programming languages used in robotic projects and (b) context in which BTs were used by respondents, disaggregated by work domain.}}%
	\label{fig:respondents}
	
		\vspace{-3mm}
	
\end{figure}

The majority of the survey respondents were from industry with 67.6\% of respondents (23/34), while academic respondents made 29.4\% of respondents (10/34). Only one respondent did not provide their work domain (2.9\% of respondents). 
The practitioners had diverse backgrounds regarding the amount of years they worked in robotics and the programming languages they used in their projects.
The majority of our respondents had between 2-10 years of experience in robotics  (55.9\% of total answers), followed by more than 10 years of experience (26.5\% of total answers) and only 17.6\% of total respondents had less than 2 years. 
\Cref{fig:respondents}, part b, shows that among respondents the most used programming languages in their robotic projects were Python (82.4\% of total answers) and C++ (79.4\% of total answers). Other languages included Rust, RAPID, KRL, and Lua (only one or two selections per language). This usage pattern align with the known fact about Python and C++ being the most popular programming languages in robotics \citep{garcia2020robotics}.

\Cref{fig:respondents} , part a, shows an overview of the context usage of BTs disaggregated by respondents' work domain. For industry respondents, the majority of context usage was in their industrial work (48.6\% of answers within the category), and less usage in academic projects. Similarly, the majority of academic respondents used BTs in their academic projects (83.3\% of of answers within the category), but no one reported using BTs in industrial projects. BTs usage in toy or tutorial projects were not popular among academic and industrial respondents (16.7\% and 27\% of answers within a category, respectively), which shows that the respondents had hands-on experience with BTs. We did not explore further the nature of the academic or industrial projects since it was out of the scope of this study.

\section{Threats to Validity}

\textbf{Internal validity:} 
As active participants in the action-research cycles, the visiting researchers may have influenced practitioner experiences and observations. We partially mitigated that through evaluation workshops at the end of each iterations where participants evaluated the observations of the visiting researchers, and had the opportunity to correct misinterpretations.

We designed our survey instrument based on the emerging action-research interest ares. 
However, we might have introduced framing bias in how questions were posed.
To reduce this, we allowed open-ended questions and "other" option with elaboration, enabling practitioners to respond in their own terms rather than being constrained by our framing. 

To ensure response quality, we implemented two control questions in different stages of the survey to identify respondents with no BT experience, whose responses were excluded. In addition, we discarded answers that did not directly answer a question during the data analysis.

Our survey focuses on exploratory and confirmatory qualitative and quantitative data. Thus, we only use descriptive statistics for the quantitative data with no hypothesis formulation \citep{stol2018abc}. To reduce bias in qualitative and thematic coding analysis, we involved two researchers. One did the coding, followed by a session with another researcher where all codes and themes were reviewed, discussing emerging codes and resolving any disagreements.

Our survey results might be biased due to our data collection method. We mitigated this risk by publishing the survey on two social platforms used by robotic practitioners, ROS discourse and LinkedIn.

\textbf{External validity: }
We have only considered BTs adoption in ROS-based systems, while there might be practitioners adopting BTs in other robotics' middlewares and frameworks. We acknowledge that limiting our study to ROS-based systems resulted in capturing ROS-experienced practitioners and missing other practitioners' experiences and practices when adopting BTs. However, we wanted to focus on the dominant middleware in robotic community that is open-source. Practitioners working with proprietary or non-ROS middleware may face different adoption experiences not captured in this study.

Action-research studies provide depth and contextual grounding.
However, the generalizability of action-research findings beyond the studied context is inherently limited. To partially mitigate that and provide breadth, we expanded the study scope and conducted the survey. 
Potential bias may persist because the interest areas identified within a single company may not cover the full range of BTs adoption concerns relevant to other contexts.
We do not claim that our findings are representative nor generalizable to all practitioners using BTs in robotics. Our goal is to shed light and build knowledge on under investigated area in BT research.

\textbf{Construct validity:} The formulation of survey questions may introduce ambiguity and fail to accurately capture the intended constructs of practices, experiences, and design decision factors. To mitigate such a threat, we piloted our survey with three experts from different organizations and companies prior to distribution. The pilot allowed us to refine 
and reduce misinterpretation of questions before the public distribution.

\textbf{Transparency and Privacy:} To ensure transparency, we provided an online appendix containing the survey instrument, and anonymized dataset as a replication package. Due to privacy, we could not provide the company notes. To ensure survey respondents privacy and consent, we collected minimum information about participants and limited access to data. We also added an introductory paragraph in the survey to disclose the study purpose, anonymity assurance, data retention notice, and right to withdraw. 
Survey participants could opt-out at any stage. 

\begin{figure}
	\begin{center}
		\includegraphics[
		width=\linewidth
		]{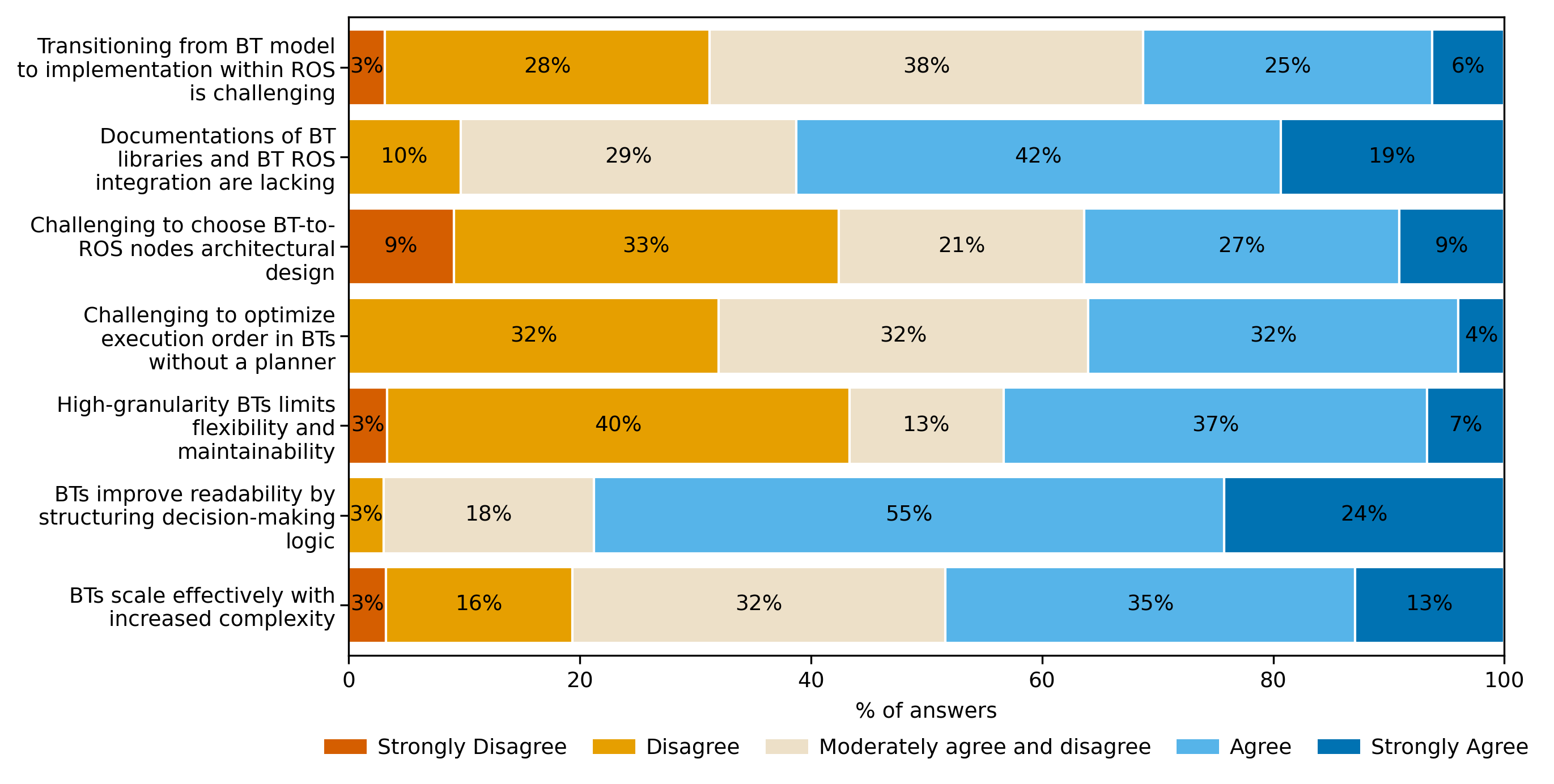}
	\end{center}
	\caption[Statements agreement]{\footnotesize{Survey responses regarding agreement levels with predefined statements from the TAR findings.}}%
	\label{fig:predefinedChallenges}
	
	\vspace{-3mm}
\end{figure}

\section{Results}
\label{sec:results}
We present the results of the  different iterations of the action-research study and the survey data. Throughout the results, we use the terms "participants" to refer to the action-research participants, "respondents" to refer to survey respondents, and "practitioners" to refer to both. We use "ID" followed by a number to refer to a respondent given ID (check the online appendix data for number reference), and "P" followed by a number to refer to an action-research participant (check \cref{subsub:studycontext} for number reference). 

The results span eight interest areas that emerged during the action-research study and enriched by the survey data: granularity of BT model, BT libraries, BT-to-ROS nodes architectural design, mixing programming languages during the programming of BTs, team communication, understandability, the usage of planning algorithms with BTs and the scalability of BTs.  

In the following, the results are organized according to the identified interest areas to ensure a coherent structure within each area. Results for these interest areas together answer RQ1-3. Each interest area is discussed in relation to the applicable research questions. Not all three research questions are necessarily covered in every section, as some interest areas were only relevant to practices, experiences or influencing factors, and we decided to scope the results accordingly.

\Cref{fig:predefinedChallenges} shows an overview of the agreement levels of survey respondents with findings regarding benefits and challenges from the action-research participants, discussed further below for each corresponding section. 
Note that percentages may not sum to 100\% due to rounding. We highlight and discuss the findings for each area.

\begin{figure}
	\begin{center}
		\includegraphics[
		width=\linewidth,
		trim=150mm 170mm 160mm 150mm,
		clip
		]{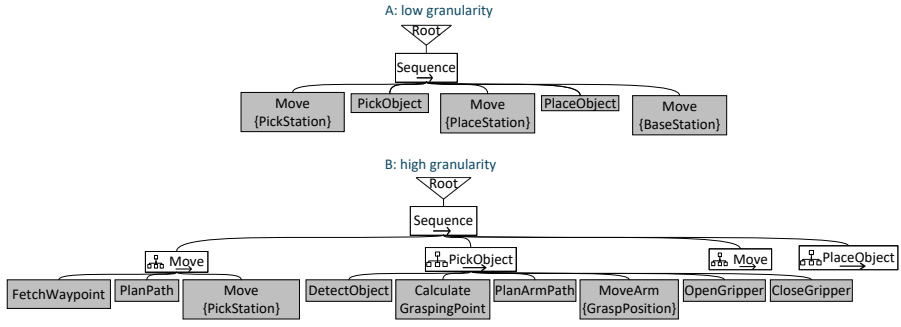}
	\end{center}
	\caption[Granularity level]{\footnotesize An example of a pick and place mission represented in two-levels of granularity in BTs (A) a low-level of granularity and (B) a high-level of granularity.}%
	\label{fig:granualityExample}
	
	\vspace{-3mm}
	
\end{figure}

\subsection{The Granularity Level of BT Model.}\label{sec:results-granularity}
By the granularity level of BT model, we refer to determining which behaviors should be represented within BT model and which should remain in the underlying code.
In the following, we present practitioners preferred level of granularity, the experiences they faced, and the influencing factors for choosing specific model design.

\subsubsection{The Granularity Level: Practices (RQ1)} 
\label{subsubsec:granularitypractice}

During the action-research study, participants were faced with a decision of what type of information from the robotic mission to represent in BT model and what details to encapsulate in the coding. To demonstrate the idea, the examples in \cref{fig:granualityExample} show a similar pick and place mission to the one in the action-research study. The mission could be modeled in BTs in two ways: (A.) at a low level of granularity and (B.) at a high level of granularity. Details such as \code{FetchWayPoint} and \code{PlanPath} could be cooperated in the code of \code{Move}, or abstracted to the BT model.
The participants decided to go with a level of detail in BTs allowing them to reuse existing code. In addition, they did not want to abstract details, such as the robotic arm path planning, into the BT model. The chosen level resembled A in \cref{fig:granualityExample}.

\begin{figure}[b]
	\begin{center}
		\centering
		\includegraphics[width=0.9\linewidth]{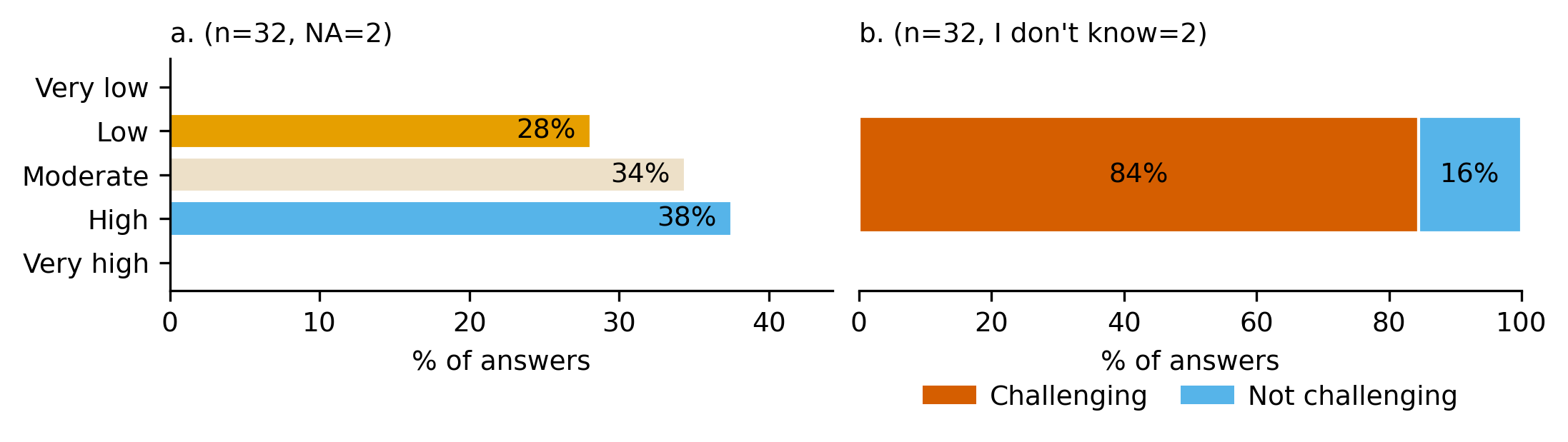}
		
	\end{center}
	\caption[Granularity of BTs]{\footnotesize{Survey responses regarding (a) the used granularity levels and (b)  the challenge of deciding the granularity level of BTs. n refers to the number of total answers, and NA refers to not applicable choice.}}%
	\label{fig:granularitycombined}
	\vspace{-3mm}
\end{figure}

\Cref{fig:granularitycombined}, part a, shows an overview of the granularity level survey respondents tend to use when modeling using BTs. No single level of granularity emerged as a majority preference. The respondents had a slight tendency toward higher granularity (38\% of total answers), followed by moderate and lower levels of granularity (34\% and 28\% of total answers, respectively). 

\subsubsection{The Granularity Level: Experiences (RQ2)} 
Determining the level of granularity when modeling a robotic mission in BTs can be challenging.  During the action-research study, we noticed that modeling a scenario with few branching possibilities, resulted in a branching explosion in the corresponding BT. We had to take decisions about the level of granularity modeled to keep the tree useful and manageable since the participants noticed an overly detailed BTs limited the flexibility to modify the tree and maintain it.

\textit{P2: a threshold for modeling flexible behavior using BTs is reached very fast. Namely, modeling even a short nominal sequence with few branching possibilities, results in a branching explosion in the corresponding BT.}

The majority of the survey respondents, 84.4\% of answers, also reported that deciding the level of modeling granularity is challenging (see \cref{fig:granularitycombined} part b). When we asked about the negative effect of an overly detailed BTs on the model modification flexibility and maintainability, the responses were unsettled. \Cref{fig:predefinedChallenges}, third raw, shows the agreement-levels of survey responses with the later challenge. 
The responses were the most near-even split responses between total agreement and disagreement levels compared to other statements in the figure (43\% of responses for each level). Only 13\% selected the neutral option, which represents the lowest neutral rate among all statements. This near-even split, combined with the low neutral proportion, indicates that the experiences of respondents with high granularity  and its negative effect on modification flexibility and maintainability vary significantly. This was apparent in the practice part with an unsettled clear favor of this design consideration (overly-detailed vs low-detailed BTs). 

\begin{finding}{Challenging to decide BTs granularity}
	Determining the appropriate granularity level of BTs was a widely recognized challenge among practitioners. However, practitioners held mixed views on 
	the extent to which high granularity negatively affects maintainability and modification flexibility.
\end{finding}

\subsubsection{The Granularity Level: Influencing Factors (RQ3)}
For elaborating on the influencing factors for choosing a level of granularity, we got 26 survey responses, plus one discarded answer for irrelevancy. We found 10 emerging factors that respondents expressed they influenced their granularity choice. Many respondents mentioned more than one factor in their response, reflecting the multifaceted nature of deciding the level of granularity. The top three factors are: (rank 1) 1. simplicity of BTs and 2. reuse of nodes, which share the same frequency, (rank 2) 3. the type of behavior or application the tree model, and (rank 3) 4. alignment with external software interfaces or requirements, and 5. the need for documentation and understandability, with the latter two sharing the same rank. Among other factors are 6. composing a modular system, 7. details preference, 8. maintainability, 9. separation of concerns, and 10. the experience in modeling.

\textit{ID14: Granularity was driven by task complexity, reusability, debugging ease, performance needs, and team structure.}

The factor "simplicity of BTs" refers to the level of granularity that maintains simplicity according to respondents' perceptions, as there is no existing metric to define a simple behavior-tree model.
The simplicity group reported that they experienced unnecessary complexity with high-granularity levels in BTs. Respondents highlighted different drawbacks of the complexity, including fragmented model, unnecessary nodes, increased information exchange between nodes, and additional boilerplate code for tasks could be handled more directly in code instead of being standalone nodes. 
Respondents that expressed both simplicity and reusability as factors experienced that finer details can be beneficial when it supports reuse of nodes and enables modular construction of BTs, as long as it does not become its own source of complexity, 

\textit{ID21: high granularity tends to fragment and obfuscate rather than enable readability, so it doesn't feel worth it. Instead I find it better to keep the number of nodes low and at a high level. There is definitely merit to higher granularity in terms of functions to reduce code duplication, but those themselves do not need to be nodes in and of themselves.}

Respondents who highlighted understandability and documentation as influencing factor mentioned the granularity level should support personal comprehension and that of stakeholders, as well as to facilitate clear documentation and communication of system behavior.
Others highlighted that granularity is context dependent, suggesting that additional detail should only be introduced when clearly required by the behavior and application type, or aligning with external software interfaces or requirements.
By behavior and application type, the respondents referred to the role of the tree (coordinated high-level tasks or executed low-level tasks), and the maturity and safety-critical nature of the robotic system. Aligning with external software interfaces or requirements meant that integrating with external interfaces and project requirements influenced the level of detail modeled in the BTs.

\textit{ID2: At first I used a very high granularity. Then I realized this was leading to unnecessary nodes. [...] I now operate at the granularity provided by the interfaces closest to the functionality. So if I am using Nav2, and I want to NavigateToPose with Nav2, I make sure the BT action talks to NavigateToPose directly. The higher-granularity alternative is having some server 'MoveAroundMap' which then calls NavigateToPose itself multiple times.}

In general, there is no one factor that influences the choice of granularity modeled in BTs. From our experience in the action-research study and the answers of respondents, we noticed it is hard to know when to stop adding details from the beginning. We observed that trial and error approach is needed, as well as the reported factors by respondents.

\textit{ID32: [...] I try to maintain the bt paradigm as far as practical, avoiding lower levels of traditional nested conditional statements. It's not been easy to always sat where "to draw the line".}

\begin{finding}{A trial-and-error mindset for BTs granularity}
	Practitioners design decisions regarding BT granularity had a multifaceted nature. Practitioners reported a trial-and-error mindset needed when approaching the decision.
\end{finding}

\subsection{The Usage of BT libraries.} 

In the following, we present practitioners' practices in terms of used BT libraries, and their experiences about BT libraries' setup, documentations and implementation.

\subsubsection{BT libraries: Practices (RQ1)}

At the beginning of the action-research study, we presented to the participants both \BTCPP and \pytrees as options to implement. The participants favored the usage of \BTCPP due to the existence of a GUI, Groot. 

\begin{figure}[!tbp]
	\begin{center}
		\centering
		\includegraphics[width=\linewidth]{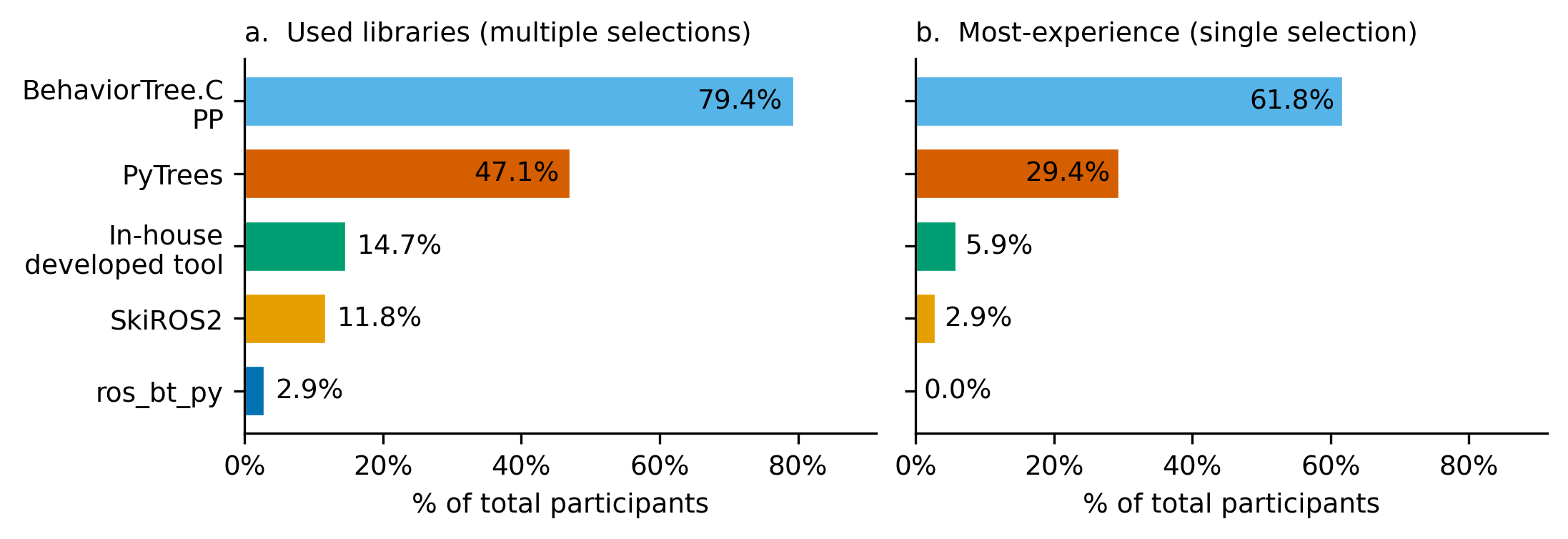}
		
	\end{center}
	\caption[Used BTs libraries]{\footnotesize{Survey responses regarding (a) the most used BTs libraries and (b) the library that participants had the most experience with.}}%
	\label{fig:BtTools}
	\vspace{-3mm}
\end{figure}

\Cref{fig:BtTools} shows an overview of (a) the BT libraries that survey respondents tend to use, where they could select multiple options, and (b) the BT library that participants had the most experience with it, one option only allowed. The presented percentages in A do not sum to 100 since multiple selections were allowed. 

The majority of respondents tend to use \BTCPP (79.4\% of total participants), and it is also the one they had the most experience with it (67.8\% of answers). 
\pytrees came next as the most used library, with just under half the percentage of \BTCPP (29.4\% of answers). Unlike the most experience data, the usage data for \pytrees shows a more moderate difference compared to \BTCPP (47.1\% of total participants). 

In-house developed tools, \skiros, and ros\_bt\_py\footnote{\MYhref{https://github.com/fzi-forschungszentrum-informatik/ros_bt_py}{fzi-forschungszentrum-informatik/ros\_bt\_py}}
 had smaller shares in the usage data (14.7\%, 11.8\% and 2.9\% of total participants, respectively). 
For the in-house developed tools, only one of the five answers provided information on the tool. The response reported the usage of the BT library that was developed by Magazino GmbH through an EU project \citep{tenorth2022controlling} and the respondent chose it as the one they had  the most experience with.
For the  experience data, the cumulative percentage of these three libraries was small compared to \BTCPP and \pytrees. Only two responses were received for the in-house developed tools, one response for \skiros, and zero for ros\_bt\_py, showing respondents had the least experience in these libraries.

\subsubsection{BT libraries: Experiences (RQ2)}

During the setup in the action-research study, participants faced positive and challenging aspects with BT libraries. 
They appreciated \BTCPP library for the ease to start visualizing the tree model using its GUI, Groot, and Groot decoupling from the implementation. However, it was hard to start implementing (coding) the tree.
They encountered multiple challenges for installing \BTCPP and connecting it to ROS 2.  
There were missing pre-requirements in the documentation of the library during the installation. We managed to mitigate this challenge by Googling errors and installing missing requirements.
\BTCPP provides wrappers to use it with ROS 2, yet the setup process was unclear at some stages. 
Following the documentation alone to setup the library and connecting it to ROS 2 was not enough. We needed to check external tutorials to understand the process and check the source code of \BTCPP ROS 2 integration.

To check if other practitioners faced similar challenge with BT libraries documentation regarding ROS integration, we collected responses regarding agreement levels. We did not specify a library in the challenge statement, rather kept it generic. \Cref{fig:predefinedChallenges}, the sixth row, shows an overview of the agreement levels.
The total agreement levels were 61\% of responses, the second highest total agreement levels compared to other statements. Only 10\% of responses disagreed, and no response strongly disagreed. The remaining 29\% of responses selected a neutral position. The minimal disagreement, combined with a relatively high rate of total agreement is consistent with the action-research participants experience.

To examine the experiences of other practitioners regarding setting up behavior-tree libraries, we surveyed respondents focusing on documentation completeness and the ease of documentation-guided setup in their projects. We also asked them to elaborate on their usage experience in general. We focused on the two most used libraries in robotics (\BTCPP and \pytrees), the academically-developed library \skiros, and provided two options for filling information about in-house developed libraries and "other". Respondents were surveyed only regarding the library with which they had the most experience. 

Due to the limited number of responses for the in-house developed libraries (n = 2) and \skiros (n = 1), there is no collective meaningful overall trend. In the following, we report the responses on a per-library basis, reporting levels of agreement and providing a summary of the elaborated responses to experiences with each library.

For the BT library developed by Magazino GmbH, the response was positive about documentation completeness and the ease of documentation-guided setup. For the other reported in-house developed library, the response highlighted a need for improving the documentation for setting up the library, and it was on the edge about documentation completeness. The response elaborated that the library performed well in specific cases but struggled outside them. Finally, the response regarding \skiros highlighted a need for improving documentation in terms of completeness and guided setup. The response indicated that \skiros enabled modular skill composition but came with a steep learning curve and runtime debugging challenges.

\textit{ID14: In my experience with SkiROS2, I found the platform to be a robust tool for developing complex robotic behaviors through skill composition. The framework's modularity and integration with ROS facilitated the organization and execution of various robot skills. However, I found the learning curve of the library to be high, particularly due to the extensive use of strings in skill descriptions. This reliance on string-based configurations, combined with Python's dynamic typing, sometimes led to errors surfacing only during runtime, complicating the debugging process.}

\begin{figure}[!tbp]
	\begin{center}
		\includegraphics[
		width=\linewidth
		]{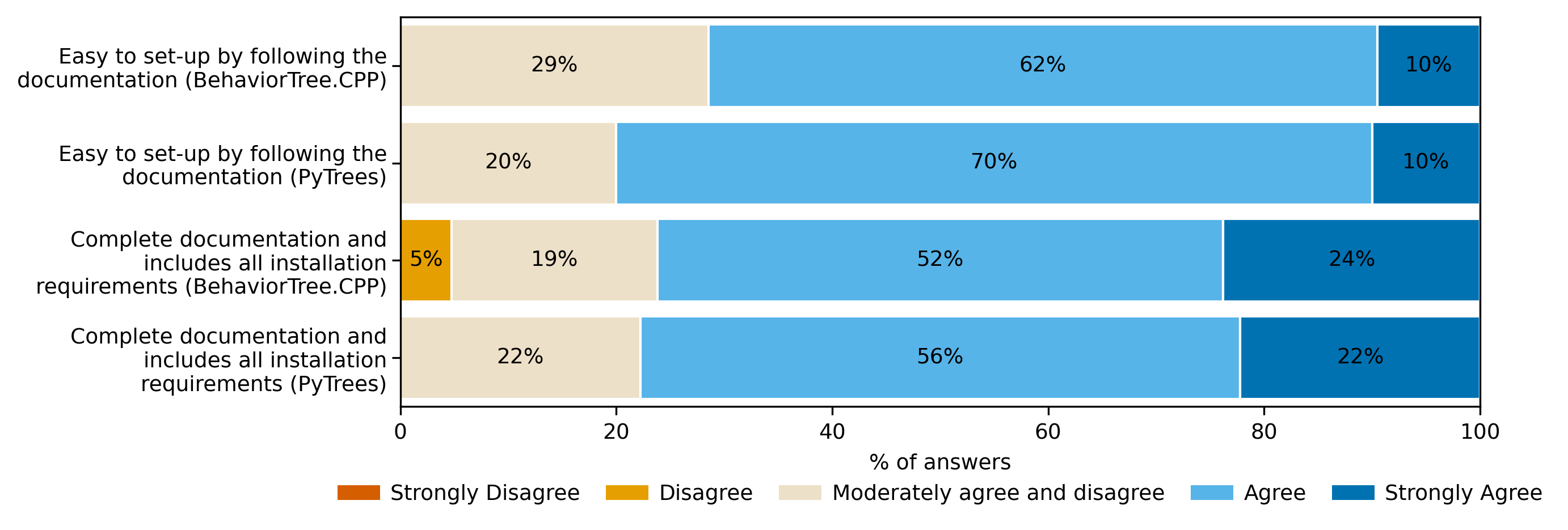}
	\end{center}
	\caption[Tools set up and documentation]{\footnotesize{Survey responses regarding documentation completeness and documentation-guided setup ease, specifically for PyTrees and BehaviorTree.CPP.}}%
	\label{fig:BTSetup}
	
	\vspace{-3mm}
	
\end{figure}

\BTCPP received the majority of responses since respondents had the most experience with it and \pytrees came next in rank (see b in \cref{fig:BtTools}), thus we highlighted further their results compared to the former libraries. 
\Cref{fig:BTSetup} shows an overview of agreement levels regarding the documentation completeness and ease of documentation-guided setup for both libraries. In terms of documentation completeness, the two libraries scored highly and were slightly comparable in total agreement (agree \& strongly agree), with only 3\% difference in favor of \pytrees. However, only \BTCPP received one negative response. Regarding documentation-guided setup, both libraries again scored high in total agreement, with a 10\% difference in favor of \pytrees.

For the elaborated responses about the usage experience, we received a total of 22 answers for \BTCPP and \pytrees. We identified three emerging codes in the answers. Based on frequency order, the codes are positive experience, implementation challenges, and comparison. Some responses addressed multiple themes, therefore applicable responses were segmented and each part coded under its respective thematic code independently.

\textbf{Positive Experiences.} Under the code "positive experience", 2 sub-codes emerged from a total of 19 responses (\BTCPP 12 responses, \pytrees 7 responses). By count order functional was the most common, followed by ease of use. Under "functional", responses for both libraries broadly expressed general satisfaction with the capabilities of the libraries. Under "ease of use", responses regarding \pytrees described the library as lightweight and easy to use, while \BTCPP responses tended to reference straightforward workflows. 

\textit{ID15 regarding \pytrees: The Python library is quite easy to use. Mostly due to Python itself. The abstract interface class behavior is simple and easy to implement. And for more complex behaviors there is more to use like the initialize and terminate methods or setup for complex setups and argument parsing. The blackboard is also easy to use and can easily store complex data structures. The composite interface is also simple so custom control nodes can be defined.}

\begin{finding}{Documentation-guided setup and integration incomplete}
	\BTCPP and \pytrees were well-received by practitioners for their functionality and ease of use. Although both libraries received high ratings for documentation completeness, practitioners reported that official documentation alone was insufficient for setup and integration with ROS, highlighting an area for improvement.

\end{finding}

\textbf{Implementation Challenges.} Under the code "implementation challenges", four sub-codes emerged from a total of 11 responses (\BTCPP seven responses, four \pytrees responses). The sub-codes are incomplete documentation, integration and low-level execution leakage, logic creation and reuse, and finally tooling, debugging and maintenance gaps. We see the emerged sub-codes as identified challenges, which this community can contribute to improving them. In the following, we elaborate on the challenges.  

\textbf{1. Incomplete Documentation.} 
Responses regarding \BTCPP highlighted the need to check source code to understand properly and the existence of hidden functionalities that are not documented. The response regarding \pytrees mentioned the need to read tutorials and watch videos to understand the best practices. 

\textit{ID28 regarding \BTCPP: The documentation is quite good, but some things still needed to be checked from their source codes to properly understand how they work.}

\textbf{2. Integration and Low-Level Execution Leakage.} 
Only \BTCPP had responses under this code. Responses regarding \BTCPP complained about the need to reason and manage low-level execution details while building a BT. A response described the need to think about low-level execution details while building BTs such as linking services and actions, and synchronous or asynchronous type of node. Another response highlighted that integrating the library with long running background tasks was sometimes error prone, without providing details about the type of tasks.

\textbf{3. Logic Creation and Reuse.} 
For \BTCPP, responses reported a difficulty to manually create and extend the XML files, a high effort for adding new leaf nodes, and several modeling shortcuts. For \pytrees, a response commented on reuse difficulty due to the underlying programming language, Python.

\textit{ID15 regarding \pytrees: The only "hard" thing is creating the BT in the script format. I created a method for example so i can reuse full subtrees when creating the BT because in Python all objects are pointers and cannot be used multiple times in the tree.}

\textbf{4. Tooling, Debugging and Maintenance Gaps.} 
\pytrees was criticized for not being actively maintained, and the lack of custom tools for supporting the library. A response outlined that debugging in \BTCPP was not easy for them outside the development mode.

\textit{ID16: PyTrees is very lightweight and easy to use, but this means you have to build custom tooling around it for your application. [...] it's not very actively maintained otherwise.}

\textbf{Comparison.} Regarding the final code "comparison", only two responses directly compared \pytrees and \BTCPP. A response expressed a preference for \pytrees, citing the ability to create BTs in regular code as opposed to being constrained to a static XML representation in \BTCPP. The response lacked clarity since as far as we know \BTCPP allows direct creation of the tree within the code. The other response noted that \pytrees was less clear than \BTCPP without elaborating on the reasons, and the practitioner continued using it due to greater familiarity with Python. These responses indicate that library usage is influenced by familiarity with programming languages and individual perceptions of the library workflow, although some perceptions may not accurately represent the actual capabilities of the library.

\begin{finding}{Gaps in libraries' ecosystems}
	Although practitioners expressed general satisfaction with \BTCPP and \pytrees usage, they reported challenges in managing low-level execution details, creating and reusing logic, and limited debugging support.
\end{finding}

\begin{figure}[!tbp]
	\begin{center}
		\centering
		\includegraphics[width=\linewidth]{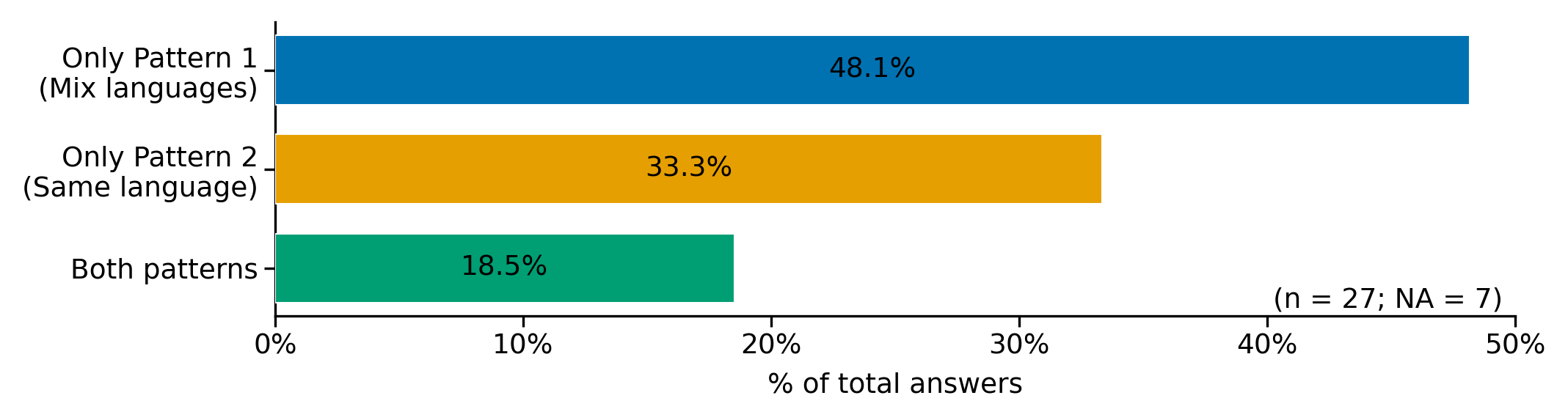}
		
	\end{center}
	\caption[Mixing programming languages in BTs]{\footnotesize{Survey responses regarding used pattern when faced with the choice to mix programming languages in BTs.  n refers to the number of total answers, and NA refers to not applicable choice.}}%
	\label{fig:mixinglanguages}
	\vspace{-3mm}
\end{figure}

\subsection{Mixing Programming Languages in BTs: Practices (RQ1)} 
\label{subsec:BTMixingLang}
By mixing programming languages in BTs, we refer to the mismatch that occurs when practitioners have pre-existing action implementations written in a programming language that differs from that of the BT library adopted for tree coordination.

In the action-research study, participants faced a decision about mixing programming languages when implementing BTs. They had  existing code for actions implemented in Rust from a previous project (move, pick, etc), whereas the behavior-tree library that was selected uses C++. This mismatch prompted a decision between supporting a dual-language setup in the same BTs, or rewriting the library actions in C++ to maintain a uniform codebase. The dual language option considered was to keep the action nodes implementation in Rust, and the coordination of the tree nodes in C++.
In this case, the participants decided to mix programming languages to avoid rewriting the tree actions from scratch and reusing other codebase from previous projects.

To examine other practitioners choices in similar situations, we asked the survey respondents about  a hypothetical scenario where they have an existing code for the tree actions (e.g. Python) and they needed to use a BT library for the tree coordination that uses a different language (e.g. C++). 
Multiple choices were given: (pattern 1) the tree in one programming language and the existing code for the tree actions in another language, or (pattern 2) start from scratch and program the actions in the same language as the BTs library. In addition, we allowed free-text for other patterns. 
We allowed respondents to select multiple options to capture all relevant usages.

\Cref{fig:mixinglanguages} shows an overview of the respondents chosen pattern. We got 27 answers from the total respondents to the survey.
The majority of survey respondents chose only pattern 1 as a preferred choice, 48.1\% of total answers. While those who chose only pattern 2 were 33.3\% of total answers. Only 5 respondents chose both patterns as preferred choices (18.5\% of total answers).
Three respondents provided an explanation for their choices which was mainly influenced by using the ROS system and the flexibility of ROS in supporting language-agnostic components.  

\textit{P3: Depends. If in ROS, I would apply Pattern 1 and keep two different languages, at first. If not ROS, I would try to maintain same language.}

\begin{figure}[t]
	\begin{center}
		\includegraphics[
		width=\linewidth
		]{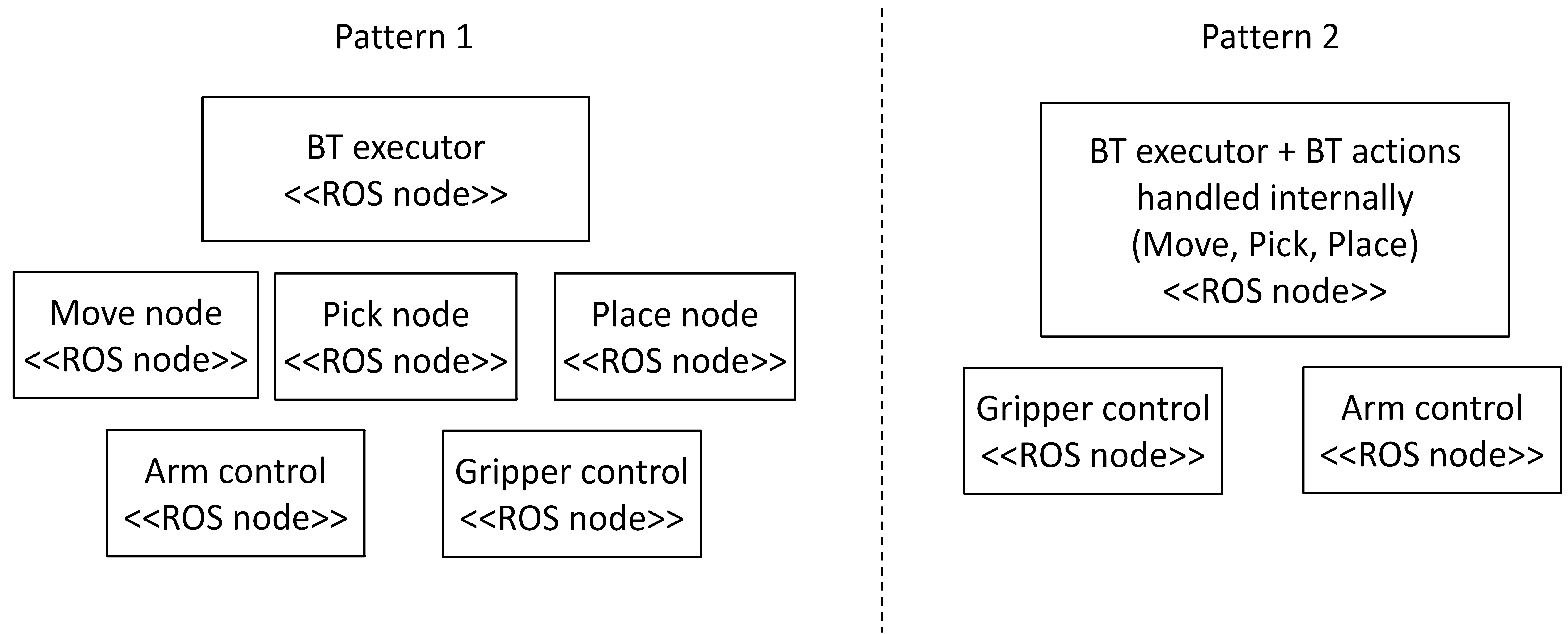}
	\end{center}
	\caption[ROS Patterns Example]{\footnotesize Conceptual illustration of the two BT-to-ROS nodes architectural design patterns observed in practice.}%
	\label{fig:RosPatternExample}
	
	\vspace{-3mm}
	
\end{figure}

\subsection{BT-to-ROS Nodes Architectural Design \& Implementation}
\label{subsec:ROSBTArch}
By BT-to-ROS nodes architectural design, we refer to how the BT execution logic and its action nodes are structurally mapped onto the ROS node graph.
During the action-research study, we observed two patterns practitioners tend to follow to deploy BTs within ROS architecture from tutorials and libraries' documentations. 

Practitioners vary in defining the boundary between the BT's internal structure and the surrounding ROS architecture. 
\Cref{fig:RosPatternExample} illustrates conceptual ROS node architectures for the two patterns. In Pattern 1, practitioners exposed each BT action node as a distinct ROS node, making the BT's decomposition architecturally visible within ROS nodes. In Pattern 2, practitioners encapsulated the entire BT execution, including all action nodes, within a single ROS node, thereby treating the BT as an internal implementation detail within a ROS node.

In the following, we present the architectural patterns that practitioners used, their experiences and the influencing factors for choosing a specific architectural design.

\subsubsection{BT-to-ROS Nodes Architecture: Practices (RQ1)} 
During the action-research study, participants reported that modeling with BTs was easy to understand and conduct on a whiteboard or using a GUI tool (Groot, in our case). However, transitioning from modeling to implementation (coding) was challenging.
From a software system architectural perspective, we could not find established guidelines on how to architect a robotic software system with BTs deployed within ROS for optimal execution. We did not find established guidelines to share with the participants to describe whether a BT should run as a single ROS 2 node, or whether each tree node should correspond to separate ROS 2 nodes. The former apply to both ROS 1 and ROS 2, and disregarding if the nodes are spread across one process or distributed across multiple processes. Participants investigated examples in the documentation of \BTCPP to understand where to place the tree within ROS and decided on a solution resembling pattern 2 (see \cref{fig:RosPatternExample}). The decision to reuse existing code (RUST code) significantly influenced the architectural design of the BT-to-ROS node
       
P2 decoupled behavior coordination from tree action implementations by distributing these functions across two ROS 2 nodes. Behavior coordination was implemented in a C++ ROS 2 node utilizing \BTCPP to execute the full BT and acting as ROS 2 action client. The tree action implementations (move, pick, etc.) were implemented in a RUST ROS 2 node acting as a ROS 2 action server. The cross-language boundary was bridged via ROS 2 action-based inter-process communication (IPC), and data exchange between tree nodes was handled internally through \BTCPP's parameterized port mechanism.
When a tree node was ticked, it sent a ROS 2 action request to the Rust node, which then invoked the relevant function for move, pick, etc. Upon function completion, the result was returned to the C++ node and mapped to a BT tick status (success, running, or failure), enabling the BT to retry or pursue alternative tree branches as specified by the BT model.

The above explained BT-to-ROS nodes architecture pattern does not reflect Pattern 2 exactly. The BT and its action tree nodes were encapsulated within a single C++ ROS 2 node, while the Rust ROS 2 node operated as a ROS 2 action server to facilitate implementation reuse, rather than to expose the decomposition of the BT model at the ROS node architectural level.

To investigate other practitioners experience, we asked the survey respondents. We gave them two patterns for placing the tree within the ROS architecture without specifying specific ROS version.
\Cref{fig:RosPatternExample} shows the two conceptual ROS architectures for integrating BTs that were given to the respondents. 
We clarified that for both patterns, it was disregarded if the nodes are spread across one process or distributed across multiple processes. We also allowed an option with free-text if respondents used other patterns.

\begin{figure}[!tbp]
	\begin{center}
		\centering
		\includegraphics[width=\linewidth]{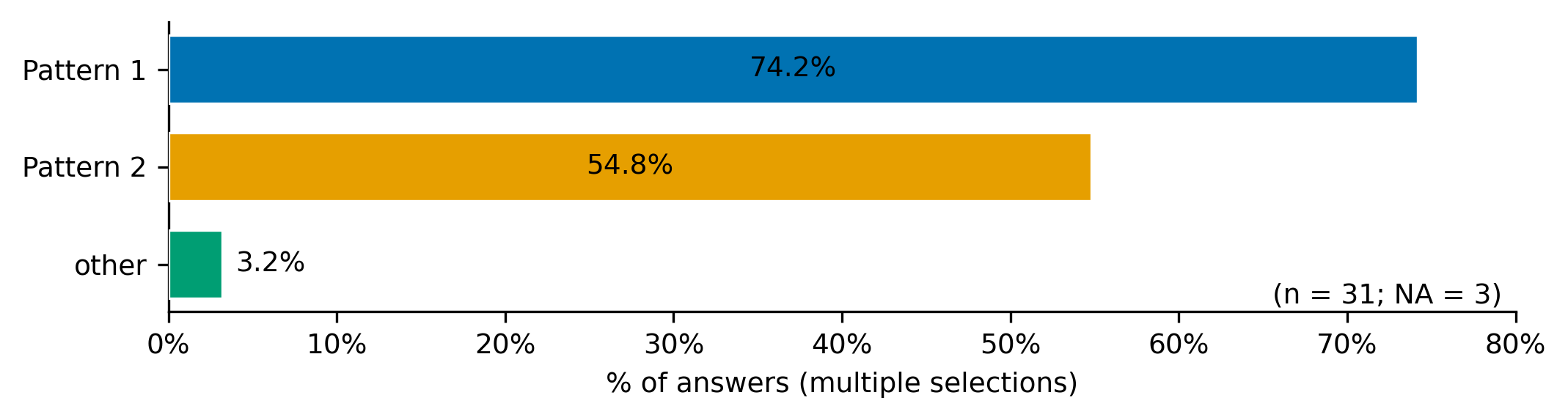}
		
	\end{center}
	\caption[Used ROS patterns]{\footnotesize{Survey responses on the BT-to-ROS nodes architectural design patterns used in practice. n refers to the number of total answers, and NA refers to not applicable choice.}}%
	\label{fig:RosPattern}
	\vspace{-3mm}
\end{figure}

\Cref{fig:RosPattern} shows an overview of the percentage of responses for each used pattern for integrating BTs within ROS architecture. 
We allowed respondents to pick multiple answers to capture the different usages. In total we got 31 answers from all respondents. The majority of respondents picked pattern 1 (74\% of answers), which was also the most chosen pattern as a single choice (13 respondents). Pattern 2 came second in choice (54.8\% of answers) and only 8 respondents picked it as a single choice. Those who picked both patterns were 9 respondents. Only one respondent gave another type of pattern they used in addition to pattern 1. The given pattern represented a functionality-based organization of behavior-tree nodes across ROS nodes, in which tree nodes associated with distinct behavioral roles are grouped and deployed within separate ROS nodes, for example navigation-related behaviors within one ROS node and manipulation-related behaviors within another.

\subsubsection{BT-to-ROS Implementation \& Architecture: Experiences (RQ2)} 

During the action-research study, two aspects emerged about BTs implementation within ROS: transition from BT modeling to implementation within ROS and deciding on the placement of BTs with the ROS node architecture.    

With no prior experience and the lack of established guidelines for BTs, the participants in the action-research study expressed that it was challenging to move from modeling to implementation within ROS. Their implementation within ROS concerns particularly referred to moving from BT design model to its ROS implementation, and managing communication between the BT and surrounding ROS components.
\Cref{fig:predefinedChallenges}, the seventh row, shows that survey responses were evenly distributed between total agreement and total disagreement levels (31\% of responses each), while 38\%, remained neutral. This balanced distribution reflects an aggregation of undecided experiences among respondents. In comparison to the action-research participants perception, the results suggest that this challenge, while real for some, is not universally encountered.

In the action-research study, participants expressed that it was challenging and time consuming to decide on how to structure BTs within ROS nodes architecture with no guidelines.
\Cref{fig:predefinedChallenges}, the fifth raw, shows the agreement levels of survey respondents with the identified challenge in the action-research study.  
The survey responses showed a slight tendency to disagree with this finding. In total, 42\% responses under total disagreement levels, whereas 36\% of responses under total agreement levels. The proportion of neutral responses was 21\%. The slight difference between the agreement-disagreement levels suggests that the challenge of deciding BT-to-ROS nodes architectural design was recognized by some practitioners, but could be hidden to others. 

\begin{finding}{Challenges of BT-to-ROS implementation \& architecture are context-dependent}
	The challenges of BT-to-ROS architecture placement and implementation were recognized among some practitioners, yet irrelevant to others, highly suggesting  that context matters.
\end{finding}

\subsubsection{BT-to-ROS Nodes Architecture: Influencing Factors (RQ3)}
In the action-research study, the design decision of ROS nodes architecture (pattern 1 vs pattern2) was influenced by participants need to reuse existing code. 
When we asked respondents about the factors influencing their choice of design, we received 20 responses. We identified 11 emerging factors in the answers of respondents. Several respondents mentioned multiple factors in their response, indicating that the decision regarding placement is influenced by multiple factors. The top three ranks based on frequency are (rank 1) 1. the ease of use, (rank 2) 2. tools used, (rank 3) and 3. separation of concerns and 4. modular design. Among other factors are 5. maintainability, 6. scale and complexity, 7. ease of understanding, 8. legacy decisions, 9. better control, 10. reusability and 11. communication overhead. 

Within the "ease of use" code, responses highlighted considerations related to simplicity and ease of prototyping, implementation, system management, and conducting architecture reviews. Additionally, a response highlighted the importance of choosing a pattern that supports smooth integration with existing workflows.

Under the code "tools used", several responses indicated that ROS version influenced pattern selection. It was mentioned that key frameworks within ROS 2 ecosystem favored Pattern 1, or made it easier to implement, such as Nav2, PlanSys2 and \BTCPP integration with ROS 2. In cases where ROS version was not explicitly referenced, decisions were associated with existing robot interfaces, or the established project architecture.

The codes "separation of concerns" and "modular design" emerged as equally prominent influencing factors. 
Responses coded under "separation of concerns" emphasized maintaining clear distinctions between responsibilities within the system architecture. Responses described organizing functionality independently, or within context-specific ROS nodes, supporting clearer allocation of responsibilities and greater control over behaviors.
Under the "modular design" code, all responses mentioned pattern 1 as preferred choice for modularity, and no information was provided regarding respondents' definitions of modularity.

\textit{ID14: Pattern 1: Chosen for modularity, distributed execution, team collaboration, and fault isolation. Pattern 2: Chosen for simplicity, lower communication overhead, and faster prototyping.}

\begin{finding}{No singular factor for architecture design decision}
	Respondents design decision regarding BT-to-ROS architecture node pattern had
	no singular factor predominant, while participants were mainly influenced by one factor, code reuse. Respondents current practices were influenced by the frameworks used and tended to favor architectures that make the system work.
\end{finding}

\subsection{The Usage of Planning Algorithms with BTs}

In BTs, the arrangement of action nodes within the tree structure encodes the ordering of tasks that constitute a robotic mission. In the majority of BT libraries, the execution priority of a tree node is determined by its left-to-right ordering. Thus, if we want to take into account all the different tasks and their fallbacks in case of failure, we should define the optimal order of the tree nodes during modeling.
When we talk about using planning algorithms with BTs, we refer to whether practitioners employed planning algorithms to automatically generate or modify the structure of BTs and the order in which tree nodes are executed, rather than manually determining them.
The order of nodes in BTs are, by nature, static, unless a planning algorithm is used to reorder nodes. 

In the following, we present results regarding practitioners use of planning algorithms with BTs, the specific algorithms employed, and their experiences in determining tree node execution order without algorithmic support.

\subsubsection{Planning Algorithms: Practices (RQ1)}

In the action-research study, participants faced multiple choices about the order of the tree nodes while modeling in BTs. They manually designed and ordered the BT nodes, relying on their own judgment to determine tasks order and overall tree structure. Algorithmic support was not utilized, as no easily implementable algorithm was identified for first-time users. 

We asked our survey respondents about whether or not they used algorithms to automatically modify BTs (generate or modify), and which algorithm they use. The majority of respondents did not use any algorithm (67.6\% of respondents). When asked about the used planning algorithms, we received 8 responses. Those who used planning algorithms did not have a clear preference on which algorithm to use.
The two planning algorithms, backward chaining \citep{colledanchise2019towards} and planning domain definition language (PDDL) based planners \citep{rico2021optimized} were mentioned by 5 respondents. No specific implementation for backward chaining or PDDL-based planners were given by respondents. 
One respondent mentioned using different algorithms such as automated planning frameworks PlanSys2, Unified-Planning (UP), and PDDLStream to use PDDL-based planners. That respondent mentioned using PlanSys2 with two PDDL planners: partial order planning forward (POPF) and temporal fast downward (TFD), and UP was used with ready PDDLs such as Tamer and local search for planning graphs (LPG). One respondent made custom algorithm that was integrated into UP.
Finally, one respondent mentioned using large-language models (LLMs) as a way to automatically generate BTs.

\subsubsection{Planning Algorithms: Experiences (RQ2)}

In the action-research study, P2 reported that deciding on an  
 optimal sequence of BTs' nodes without a planner to reorder automatically the nodes was challenging.
The participants questioned the optimality of the execution order and the reactiveness of BTs. 

\textit{P2: sequence optimality is not guaranteed that the BT will ensure the optimal action sequence}

P1 and P2 felt that even having a few fallback tasks within the robotic mission resulted in easily reaching the complexity of BT modeling.
One participant summarized the tradeoff between BTs and planning algorithms in the following:

\textit{P2: BTs provide a more straightforward and visually intuitive way to model behavior, with conditions implicitly managed through the tree structure. This approach simplifies design and is well-suited for tasks with predictable flows, making it easier to quickly set up and adjust. However, BTs are less flexible, as they depend on a fixed structure, which limits adaptability in dynamic and complex scenarios. Planning algorithms, in contrast, offer advanced flexibility and control [...] The trade-off, however, is complexity: planning algorithms require more detailed modeling and can be more resource-intensive to set up and execute. In essence, BTs are ideal for simpler, stable tasks, while planning algorithms are necessary for environments demanding high adaptability and resilience to uncertainty.}

In \Cref{fig:predefinedChallenges}, the fourth row, the survey responses were balanced and did not yield definitive conclusions concerning the challenge of optimizing the ordering of BTs' nodes in the absence of a planning algorithm. The percentages were 32\% of responses under agree, 32\% of responses under disagree, and 32\% of responses remained neutral. Only 4\% of responses were under strong agree, the lowest between all statements, and no responses under strongly disagree.

\begin{finding}{Absence of a planning algorithm challenging}
	Practitioners reported the challenge of optimizing the ordering of BTs' nodes in the absence of a planning algorithm. However, among all practitioners no definitive conclusions about the need for a planning algorithm is reached. 
\end{finding}

\subsection{BTs \& Team	Communication: Experiences (RQ2)} 
In the following, we present the experiences of practitioners regarding the effect of using BTs on teams communication. We deliberately left the nature of this effect open to practitioners' own interpretation, rather than directing them toward a specific outcome.

During the different stages of the action-research study, we observed that having the behavior-tree model eased the communication between participants. Having the model made it easier to explain the possible tasks combinations for the explored mission.

\begin{figure}[!tbp]
	\begin{center}
		\centering
		\includegraphics[width=\linewidth]{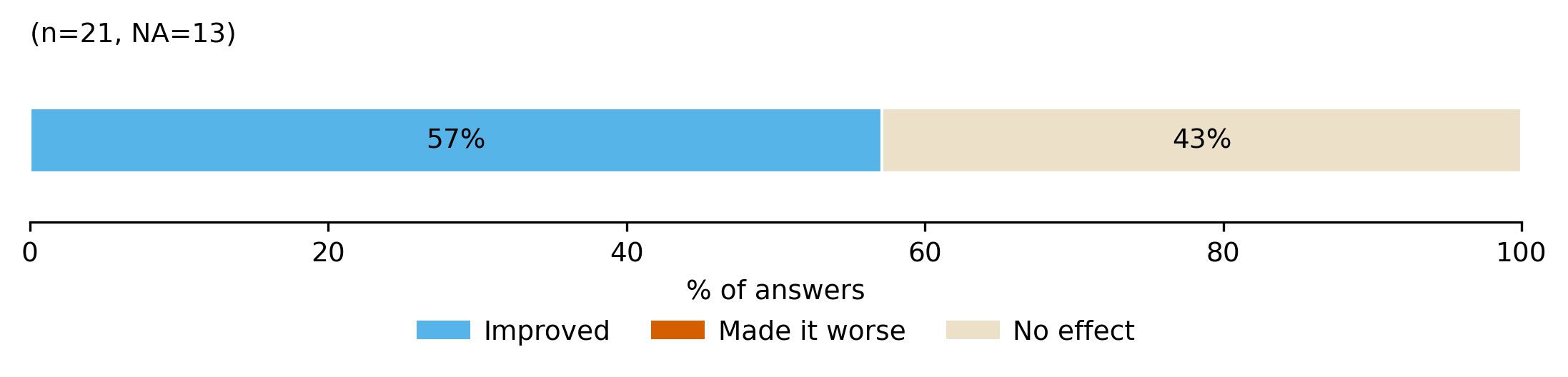}
		
	\end{center}
	\caption[Effect on Communication]{\footnotesize{Survey responses regarding the effect of using BTs on communication between stakeholders in projects. n refers to the number of total answers, and NA refers to not applicable choice.}}%
	\label{fig:communication}
	\vspace{-3mm}
\end{figure}

To understand other practitioners experiences, we asked the survey respondents about the observed effect of using BTs on communication between stakeholders in their projects, with 13 responses to this question, and other survey respondents reporting the question was not applicable to their work. \Cref{fig:communication} shows an overview of the answers. The majority of respondents noticed an improvement in communication (57\% of answers), while 43\% did not notice any difference. No one reported that the communication got worse due to using BTs.

When we asked respondents to elaborate further about the effect on communication, the majority reported that BTs made it easier to understand the robot behavior and communicate technical details with developers or even less technical colleagues (61.5\% of answers), which was similar to our action-research experience. One respondent expressed that BTs made it efficient to do changes and adapt different functions during the project life span. Another respondent noticed  improvement in the collaboration between stakeholders.  

\textit{ID12: Similar to forms of Model Based System Engineering, the use of BTs simplified the ability to talk through technical details, as well as speed the ability to do risk management, change impact analysis, and other related functions over the life of the system}

Survey results found that communication was affected depending on the role of the stakeholders (23.1\% of answers). One respondent who reported a positive effect expressed that less technical stakeholders appreciated the visual representation of the mission using the behavior-tree model. Another respondent reported nearly the opposite: that it was easier to communicate with other developers (technical peers), but there was no effect on the communication with stakeholders with less technical background.

In commenting on communication benefits, survey respondents compared BTs to to state machines, a different type of behavior model, expressing that BTs improved communication compared to using state machines in other projects (38.5\% of answers). One respondent highlighted that introducing BTs in projects and convincing stakeholders to switch from their usual models was challenging. At the beginning of the action-research process, we also had similar experience. Later in the study, when making changes and communicating the technical details of the changes was made easier using BTs, the researcher started noticing a switch in the opinion of one of the study participants. 

\textit{ID1: At first it was complicated to make the stakeholders understand the benefits of BTs over other policies (e.g., FSM) but when that happened, a project where BTs were used was more positively seen than the alternatives (somewhat linked to 'higher level of AI/intelligence involved')}

\begin{finding}{BTs facilitate team communication}
	Practitioners perceived BTs to improve team communication by making it easier to understand the robot behavior and enabling clearer communication of technical details among stakeholders with varying technical expertise.  However, the extent of this benefit may vary depending on stakeholder role and familiarity with BTs.
\end{finding}

\subsection{BTs \& Understandability: Experiences (RQ2)}

In the following, we presents the results about the perceived perception of practitioners about the effect of BTs on understandability. Specifically, we investigated the effect of BTs on the readability of the decision-making logic and on the understandability of robotic missions execution.

In the action-research study, participants observed that compared to state machines, the previously underlying behavior model for used tool by the company, it was easier to understand BT model. BTs provided them with clarity regarding reading and understanding the decision-making logic. In addition, participants also found that understanding the decision-making logic, made clear by BTs, helped them early-identify and represent requirements for different tasks, represented in tree nodes.

\textit{P1: When we see the mission in BTs, then in the early phase we can add the operator requirements, the safety expert requirements, and so on. [...]  Compared to SP and state machines, BTs were easier to understand, but to start programming with it was harder.}

\Cref{fig:predefinedChallenges}, second row, shows that respondents perceived BT model as an enhancing model for reading and understanding the decision-making logic.
Compared to other statements in \cref{fig:predefinedChallenges}, this finding had the highest total agreement levels.
The overall agreement levels were 79\%, including the highest strong agree level among all statements 24\%. Only 3\% disagreed, with no strong disagreements.

\begin{figure}[!tbp]
	\begin{center}
		\centering
		\includegraphics[width=\linewidth]{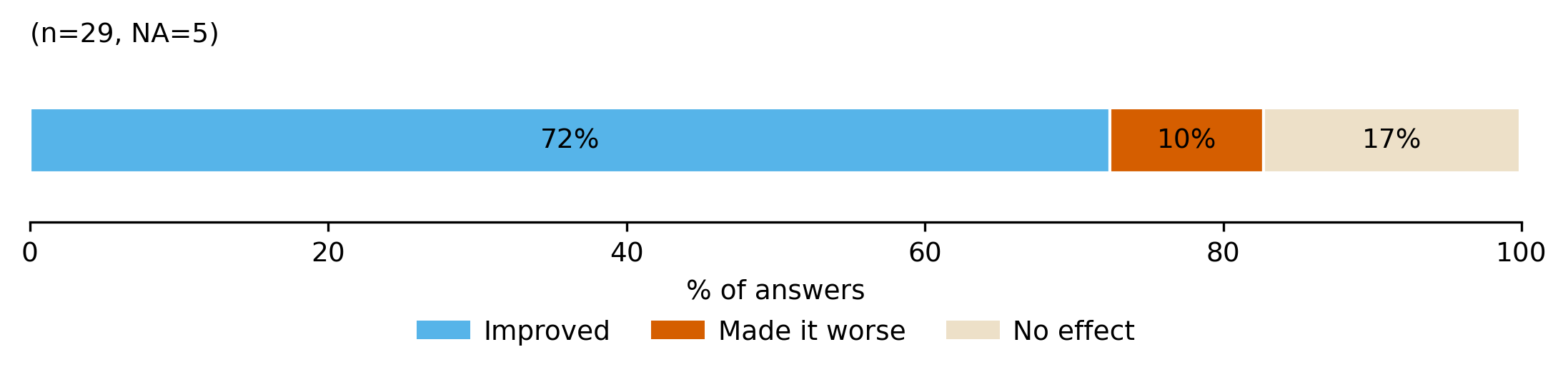}
		
	\end{center}
	\caption[Effect on understandability]{\footnotesize{Survey responses regarding the effect of using BTs on understanding the robotic missions' execution and decision-making in projects. n refers to the number of total answers, and NA refers to not applicable choice.}}%
	\label{fig:understandability}
	\vspace{-3mm}
\end{figure}

To understand better the experiences, we asked the survey respondents about the observed effect of using BTs on the understandability of the robotics missions execution and decision-making process as perceived by different stakeholders or  personal perception. \Cref{fig:understandability} provides an overview of the respondents answers. The majority of respondents reported improvement in the understandability of these aspects (72\% of answers). On the other hand three respondents reported negative effect on the understandability aspects (10\% of answers), while five respondents reported no significant positive nor negative effect on understandability (17\% of answers). 15 responses elaborated on the effect of BTs on  understandability aspects.   

Diving deeper into the reasons, three respondents felt the negative effect of BTs on understandability, was due to either comparing them to previous experience with state machines, or introducing them into new audience.
Among the respondents who compared BTs to state machines, one expressed that state machines are better for understandability, although no justification was provided. Another respondent highlighted that stakeholders were generally more familiar with state machines than with BTs.
Two respondents reported BTs required an initial effort for comprehension for stakeholders, or even themselves. They reported that stakeholders took time to understand the execution flow and ticking principles associated with BTs. However, both respondents reported that the benefits that came with BTs out-weighted the initial-comprehension effort.

\textit{ID16: I actually think that BTs are not the most intuitive tools to understand (it takes a while to rewire your brain), and visualizing large trees can sometimes be intimidating -- especially if they are too granular. However, to me, the benefits of having modular behaviors that are easy to integrate outweigh the slight drawback in understandability.}

Among respondents who reported a positive effect, the majority highlighted the positive effect of using BTs on understanding and clarifying the functionality and structure of robotics missions and systems (66.7\% of answers). Having the visual representation of missions logical flow in BTs helped respondents comprehend and clearly see the logic and reasoning behind the switching between systems functionalities. In addition, BTs helped them understand clearly the overall structure of missions. 

In general, the positive responses group emphasized that the visual representation and modular design capabilities of BTs positively influenced their understanding. A few additional highlights from them, though less common, were the eases to modify and maintain missions, ease of communication between stakeholders, and ease to think about a mission alternatives.

\textit{ID13: The modular structure of BTs allows complex behaviors to be broken down into smaller, reusable components, making the system's logic more transparent and easier to comprehend. This hierarchical organization not only clarifies the flow of execution but also facilitates debugging and maintenance by isolating specific behaviors within distinct nodes. Additionally, the visual representation of BTs provides an intuitive overview of the system's behavior, aiding both developers and stakeholders in grasping the project's functionality.}

\begin{finding}{BTs enhance understandability}
	Practitioners perceived BTs as improving the readability and understandability of robotic decision-making logic. The visual and modular structure of BTs was identified as a driver of this benefit.
\end{finding}

\subsection{The Scalability of BTs: Experiences (RQ2)}
In the following, we present the results about practitioners perception of BTs scalability as system complexity increases. We deliberately did not prescribe a specific dimension of scalability, leaving the interpretation to practitioners themselves. Looking at the results, practitioners discussed scalability along two dimensions: how the tree structure holds up as systems grow, and how well existing libraries support development at scale.

In the action-research study, participants did not try different mission sizes to evaluate the scalability of BTs. However, the scalability of BTs was discussed by participants. It was speculated by participants that BTs might scale as a project gets complex. However, it was anticipated that with larger projects the positive side of managing and visualizing trees might be affected negatively. 

\Cref{fig:predefinedChallenges}, the first row, shows that the respondents of the survey slightly agreed that BTs scale effectively as a project complexity increase.  A total agreement levels (agree \& strongly agree) of 48\% of responses, the total disagreement levels (disagree \& strongly disagree) was 19\% of responses, and 32\% were neutral. This moderate agreement and neutrality indicate a generally positive but cautious perception of BTs scalability. Although some respondents view BTs as highly scalable (13\% strongly agree), but the lack of a clear majority suggests perceptions of scalability remain unresolved.

Survey respondents experience with BTs scalability showed three different themes. Positive experience while suggesting improvements (41.2\% of responses) and negative experience while also suggesting improvements (23.5\% of responses). There was three responses that reflected elements of both the positive and negative experiences, indicating some thematic overlap (17.6\% of responses). The final theme was respondents expressing not having enough experience to comment on the topic (17.6\% of answers). 

In the positive experience group, respondents expressed that the model structure inherently support scalability. Respondents reported that the possibility to structure the behavior-tree model as subtrees, allow them to keep the model design modular and flexible. In addition, the available software libraries, \pytrees and \BTCPP, were praised for supporting scalable reuse through structural composition of subtrees, and supporting data variability for reused subtrees through parameterized interfaces (ports or blackboard bindings). However, two respondents suggested that available software libraries need to improve the ease of parametrization of tree nodes by making it intuitive and allowing under specification, especially to scale in the face of unknown runtime conditions. Finally, an improvement was suggested to provide scalability analysis of the computational overhead of ticking many nodes at high frequency in large-scaled BTs.

\textit{ID14: BTs scaled well in my experience due to their modularity and hierarchical structure. They allowed incremental growth, reuse of subtrees, and clear organization, which kept complexity manageable as the system expanded.}

The negative experience group reported that BTs scale poorly when practitioners model low-level tasks in the tree (high-level of granularity), leading to nodes explosion. Two respondents described the need for expert modeling skills for BTs to scale and remain maintainable. Three respondents found it challenging to develop and debug large-scale BTs due to limited support in available software libraries. A respondent highlighted a fallback in current software libraries that leaks data-flow handling (fake actions for data sharing or retrieval) into the behavior-tree model. Of course this can lead to nodes explosion affecting scalability of BTs. No specific library was mentioned for the aforementioned limitations. Finally, a respondent criticized that BTs scalability is limited by how reactivity is modeled in practice. To achieve reactiveness developers tend to distribute reactive conditions across multiple parts of the tree (explicit conditions). It was highlighted that as systems grow in scale, this manual placement becomes increasingly burdensome, making large-scale BTs harder to structure and maintain compared to hierarchical state machines, where transition logic is automatically inherited through the state hierarchy. Another respondent shared that the way reactivity is modeled made BTs better than other models such as state machines, but it needed improvements.

\textit{ID16: As a system scales, BTs can get tricky to use compared to hierarchical FSMs, for example. I think the HFSM model probably scales better since you can encapsulate things a lot better, whereas even if you introduce hierarchy in a BT, the way that "reactivity" works means you still have to manage being deep in one level of a tree but having to pay attention to reactive condition checks all the way on the other side of the tree. Once again, this can be managed by making sure trees are not too granular.}

From the different themes groups, seven respondents compared BTs scalability to other models. A shared opinion by four respondents was that BTs scaled better than state machines in their work. One respondent expressed that complexity grows in scaled systems regardless of the used behavior model. Another respondent experienced lower verification and validation properties in BTs when scaling compared to state machines and petri nets. 

\textit{ID5: Depends on the granularity, needs of more nodes types highlight BTs models aren't good enough, though practical. Need of "tricks" and concepts that don't belong to BT model, e.g. sharing data or actions to retrieve data which aren't truly actions but mechanisms. Properties for verification and validation are way lower compared to FSM, Petri nets etc.}

\begin{finding}{Support needed for BTs scalability}
	Practitioners perceptions of BTs scalability are cautiously positive but inconclusive. Practitioners with positive and negative experiences suggested improvements to current BTs libraries and practices, indicating the current state is not considered sufficient.
\end{finding}

\section{Discussion}
\label{sec:discussion}

We now discuss our findings along three dimensions. First, we revisit the interest areas and interpret what the combined action-research and survey evidence tells us, including a comparison to prior work. Second, we surface a cross-cutting observation that emerged across multiple interest areas and that, we argue, shapes how the individual findings should be read. Finally, we discuss implications for practitioners and  researchers.

\subsection{Interpretation of findings by interest area}
\label{subsec:findingsinterpretation}

\newcommand{\mypar}[1]{\smallskip
\noindent{}\textbf{#1}. }

\mypar{Granularity: complexity unsettled}
In software modelling, deciding the level of granularity to model is a common challenge~\citep{maier2017model}. Similar results regarding BT modelling emerged during the action-research study and the survey (see~\cref{fig:granularitycombined} part b). During the action-research study, the participants noticed that modelling fine details of a robotic mission in BTs negatively affected the ability to change and maintain the model. However, the results of the survey were mixed about the effect of granularity on the  flexibility and maintainability of BT models (see~\cref{fig:predefinedChallenges}).

Considering the influencing factors reported by respondents (\cref{sec:results-granularity}), we note that ID2's account of aligning granularity to the nearest functional interface (exemplified by connecting BT actions directly to Nav2's \texttt{NavigateToPose} rather than wrapping them in a bespoke \texttt{MoveAroundMap} server) offers a concrete heuristic that, to our knowledge, has not been articulated explicitly in the BT literature. We return to it in the practitioner implications below.

\mypar{Scalability: unresolved}
The participants of the action-research study speculated that BTs might scale effectively as project complexity increases. In the survey, the overall perception about BT scalability was mixed (see \cref{fig:predefinedChallenges}). Like any other behavior model, depending on the number of nodes and the complexity of the tree, BTs might not scale well, though they may still be useful for maintenance.

Looking more closely at the elaborated survey responses, we note that practitioners discussed scalability along two distinct dimensions: the \emph{structural} question of whether the tree model holds up as missions grow, and the \emph{ecosystem-level} question of whether available libraries support development at scale. On the first dimension, respondents highlighted that, compared to state machines in their company, BTs scaled better in the face of \textit{variability}. This  observation is consistent with García et al.'s finding that BTs are particularly useful for scaling to scenarios with many sources of variability~\citep{garcia2023software}. On the second dimension, respondents flagged concrete gaps: debugging support, parametrisation ergonomics, and the leakage of data-flow handling into the tree model. We argue that separating these two dimensions is useful going forward, as the remedies differ: the structural dimension calls for modelling discipline, whereas the ecosystem-level dimension calls for tooling investment.

\mypar{Understandability and team communication: confirmed}
The clarity in the decision-making logic that BTs provide was one of the reasons the participants in the action-research study wanted to participate and experiment with BTs. After using BTs in the action-research phases, this advantage was confirmed. The survey responses acknowledged that BTs are highly regarded as an enhancing model for understanding decision-making logic.
In addition, we observed during the action-research study that the presence of the BT model facilitated the communication between participants and provided a way to explain the possible task combinations for the explored mission. This benefit is a well-documented advantage of visual artefacts~\citep{moody2009physics}.

\mypar{Need for a planner: inconclusive}
The participants of the action-research study preferred to use a planner to optimise execution order, although they did not experiment with one for BTs. The inconclusive agreement levels in the survey indicate that respondents recognise the challenge of optimising execution order of BT nodes, and the potential need for a planner, but that it is not a major concern in the studied population.
Although no clear agreement was reached for this finding, we see an opportunity for the robotic community to research further and improve the ease of access to such planners. In the action-research study, participants highlighted that the reactivity of the BT model might get compromised, which is a highly claimed advantage of BTs in the robotic community~\citep{colledanchise2018behavior}. Reactivity can still be achieved by modelling the different options of a task and its fallbacks, but with static BTs the model can quickly reach high complexity, as experienced by  the action-research participants  during their work. 
Furthermore, the survey by \citet{ingrand2017deliberation} highlighted that relying solely on pure planners or purely reactive behavior models, which are triggered by the changes in the environment, is insufficient for robotic operation in real-world environments. There is a need for hybrid approach for robots to handle uncertainty in the environment \citep{ingrand2017deliberation}. 
In~\citet{garcia2023software,hallen2024behavior}, it was reported that companies were counting on operators to deal with uncertainty in the environment and re-plan the robotic mission. \citep{garcia2023software} emphasised the need for better libraries and tooling to deal with uncertainty. Existing planning algorithms can synthesise a tree, e.g., back-chaining~\citep{colledanchise2019towards}. The researchers and industrial partners within the EU project CONVINCE\footnote{\MYhref{https://convince-project.eu/}{convince-project.eu}} reported the challenge of having flexible and reactive BTs without a planner, and of accounting for uncertainty in the environment while modelling in BTs. They suggested using a Markovian planning algorithm in the CONVINCE tool chain to synthesise BTs and account for uncertain and dynamic environments~\citep{street2024towards}.

Thus, the struggle is not unique, and the robotic software community appears to be moving towards solutions to address it. However, as far as we know, the existing planning algorithms for BTs are not easy to implement for practitioners with limited experience.

\mypar{BT-to-ROS architectural design: uncertainty recognized}
In the action-research study, participants had no prior experience with BTs, in contrast with survey respondents, around 82\% of whom had over two years of experience in robotics. Participants struggled with deciding appropriate BT placement within ROS node architecture and integrating BTs within ROS. Meanwhile, survey respondents did not reach an agreement about struggling with making such decisions. We speculate that this challenge is due to the lack of established guidelines, and to the decision being treated as an afterthought left to one's experience within robotics. Establishing guidelines for best practices in placing BTs within the ROS system architecture is therefore necessary. \citet{malavolta2020you} provide guidelines for architecting ROS-based systems, and one of their recommendations is to \textit{``Group nodes and interfaces into cohesive sets, each with its own responsibilities and well-defined dependencies''.}, as captured in our survey Pattern 1 option. Yet more than half of our respondents (54.8\%) indicated they also use Pattern 2. 
We speculate that the modularity of Pattern 1  can introduce operational and communication overhead, which could explain why practitioners also rely on Pattern 2.

We further note a revealing divergence between our action-research participants and the survey respondents on the factors shaping this decision. In the action-research study, architectural decisions were driven primarily by a single factor: the need to reuse existing Rust code, which shaped a Pattern-2-aligned design. Survey respondents, by contrast, most commonly cited \emph{ease of use} and \emph{tools used}, with frameworks such as Nav2 and \BTCPP's ROS 2 integration, encouraging Pattern 1. We argue that this divergence is best understood as a trade-off between \emph{framework-aligned designs}, which optimise for integration with the surrounding ecosystem, and \emph{legacy-aligned designs}, which optimise for preserving prior investment. Neither is intrinsically better; which one a team should follow depends on where the project sits between a greenfield and a brownfield setting.  Existing guidelines need to be updated to account for this important context.  

\mypar{Documentation gap: confirmed}
During the action-research study, we struggled with incomplete documentation for setup and ROS integration. Compared to the survey responses, the struggle seems universal across different libraries (see~\cref{fig:BTSetup} and~\cref{fig:predefinedChallenges}). The current documentation of BT libraries, and the available ROS wikis guiding BT integration within ROS, are incomplete, and this is an actionable area for improvement for the community. The documentation gap appears to be a consistent problem in software robotics: prior empirical studies have identified documentation deficiencies as a recurring challenge in ROS-based software development~\citep{garcia2023software,canelas2024understanding}.

The reported challenges with \BTCPP and \pytrees suggest that the ecosystems surrounding these libraries have not matured at the same pace as their core functionalities, necessitating further intervention from the software development community.

\mypar{Transitioning from BT modelling to ROS implementation: undecided}
In the action-research study, participants described the transition from BT modelling to ROS implementation as a central challenge, particularly in moving from the BT design model to its ROS implementation and in managing communication between the BT and surrounding ROS components. The survey, however, produced an evenly distributed response: 31\% of responses under total agreement, 31\% under total disagreement, and 38\% neutral (\cref{fig:predefinedChallenges}). Taken together, these findings suggest that the challenge is real for some practitioners, but not widely shared across the broader community.

\subsection{Novice--experienced asymmetry}
\label{sec:discussion:noviceExperienced}

We observe a pattern runs across several findings: challenges that were salient to our action-research participants, who were new to BTs, appear less salient to the majority of survey respondents, who were familiar and experienced using BTs. This is most visible for the transition from BT modelling to ROS implementation (see \cref{subsec:findingsinterpretation}), and again, more subtly, for the BT-to-ROS architectural decision challenge.

This asymmetry admits two interpretations. It could reflect that these challenges are genuinely \emph{onboarding costs}, real and meaningful for adoption, but progressively resolved as practitioners accumulate experience. Alternatively, it could reflect that experienced practitioners have internalised workarounds such as source-code reading and in-house conventions to the point where the underlying gap is no longer perceived as a gap. Towards the former, the near-universal agreement on documentation incompleteness (\cref{fig:predefinedChallenges}, sixth row) fits: experienced practitioners would have long since filled the documentation gaps with their own mental models, yet still recognise the gaps when asked directly. Towards the latter, the fact that our respondents rate library documentation as highly complete (\cref{fig:BTSetup}) while simultaneously reporting that it is insufficient for ROS integration suggests a quiet normalisation of incompleteness.

The two interpretations have different implications. If these challenges are onboarding costs, the community's effort is best directed at entry-level resources, specifically, tutorials and worked examples. If they are normalised gaps, the effort is better directed at the underlying artefacts, such as completing the integration documentation, building first-class debugging tools, and codifying the conventions that experienced practitioners already follow. We argue that both are needed, and that disentangling the two is itself a valuable research agenda.

\subsection{Implications for practitioners}
 Our findings suggest several concrete directions for practitioners considering or refining BT adoption, and for the maintainers of BT libraries and their surrounding tooling.

First, practitioners making \emph{granularity} decisions may want to adopt ID2's interface-aligned heuristic: align the granularity of a tree node to the level of the closest functional interface already present in the system. In a ROS 2 project using Nav2, this means letting a BT action correspond directly to \texttt{NavigateToPose}, rather than wrapping it in a bespoke composite. The heuristic does not dissolve the granularity trade-off, but it offers a defensible default that reduces both trial-and-error overhead and unnecessary node proliferation. Practitioners may want to depart from it only when they have a specific reason, e.g., when a functional interface is too coarse to expose decision points that matter for fallback behavior.

Second, practitioners making \emph{architectural} decisions may want to start from framework defaults. Our results suggest that, in the absence of legacy constraints, Pattern 1 is what the current ROS 2 ecosystem encourages and supports most directly. Teams carrying legacy code, however, should not force themselves into Pattern 1; our action-research case illustrates that a Pattern-2-aligned design can be justified on code-reuse grounds. Framing the decision as a trade-off between framework alignment and legacy preservation makes the decision explicit rather than emergent. This framing also complements prior architectural guidance for ROS-based systems~\citep{malavolta2020you}, which recommends grouping nodes into cohesive sets with well-defined dependencies, by clarifying how BTs fit into that grouping.

Third, tool developers working on BT libraries could prioritise the integration and debugging aspects of the library ecosystem over the core modelling APIs, which practitioners already rate positively. Specific areas that surfaced across our data include: completing and validating ROS 2 integration documentation; reducing the leakage of low-level execution details (e.g., synchronous versus asynchronous node types, and service and action linking) into the modelling layer; and improving debugging ergonomics, particularly outside development mode. For \pytrees, respondents flagged maintenance activity as a concern; for \BTCPP, the opacity of some functionalities until source code is consulted. These are not deficiencies of the core libraries, but gaps in the surrounding ecosystem that the community can contribute to closing.

Fourth, teams considering \emph{planning algorithms} with BTs currently face a fragmented landscape: multiple algorithms such as back-chaining~\citep{colledanchise2019towards}, PDDL-based planners~\citep{rico2021optimized}, and more recently, LLM-driven generation \citep{izzo2024btgenbot,ao2025llm,zhou2024llm}, with no clear community guidance and limited accessible implementations. Teams whose missions exhibit modest branching and reasonable structural stability can often do without planner support; teams whose missions involve substantial uncertainty or dynamic replanning needs may want to track ongoing work in this space, such as the CONVINCE project's Markovian-planning approach~\citep{street2024towards}.

\subsection{Implications for researchers}

Our findings also open several directions for the research community.

First, the widely reported but variably experienced challenges of granularity, architecture, and planner need point to a need for \emph{empirically grounded guidelines} rather than prescriptive best practices. Guidelines that acknowledge the trade-offs we have named, and that link recommendations to project attributes such as legacy presence, mission complexity, and team experience, would fit the data better than one-size-fits-all prescriptions. The practical guideline proposed by \citet{dortmans2022behavior} offers a starting point, but lacks the broader practitioner validation that our data could inform.

Second, our data underscores a \emph{documentation and tooling gap} around the BT--ROS integration boundary. This gap connects to prior findings on ROS documentation shortfalls~\citep{garcia2023software,canelas2024understanding} and suggests that the BT community faces a variant of a broader robotics-software pattern. Research into systematic documentation practices, worked reference implementations, and automated integration testing for BT--ROS stacks would address a pain point reported consistently across our respondents.

Third, our results on planning algorithms with BTs connect directly to ongoing work in the CONVINCE project~\citep{street2024towards} and related efforts~\citep{colledanchise2019towards,rico2021optimized} on synthesising BTs under uncertainty. The inconclusive agreement in our survey data suggests that the community has not yet articulated the planner-versus-no-planner trade-off in terms that practitioners can apply to their own projects. Translating the benefits of automated synthesis into concrete, library-integrated capabilities with accessible documentation remains, in our view, the most valuable direction in this sub-area.

Finally, the comparison between BTs and alternative behavior models surfaced repeatedly in our data as an organising frame for practitioners' judgements. 
This involves the comparison to state machines in particular, but also Petri nets, as mentioned by one of our respondents (ID5).
Future empirical work on BTs would benefit from treating this comparison as a variable, e.g., by examining how prior state machines experience shapes practitioners' perceptions of BT understandability and scalability.
This would help disentangle which perceived properties of BTs are intrinsic and which are relative to the model from which the practitioners are migrating.
\section{Related Work}

Behavior trees have gained popularity in the last decade. The majority of existing research investigates technical and formal aspects of adopting BTs, with little attention to how practitioners adopt and work with BTs in real systems or their experiences with BT use.

The work of \citet{hallen2024behavior} and \citet{dortmans2022behavior} presented lessons learned from applying BTs and proposed a guideline to support BT development in practice, respectively. In \citep{hallen2024behavior}, authors reported lessons learned, identified challenges encountered, and offered recommendations to mitigate these challenges in a case study on the application of BTs in an industrial context.
Although valuable insights were presented, the authors reported their experiences only from an implementation perspective  for the reported use case.
In \citep{dortmans2022behavior}, the authors addressed the lack of guidelines for implementing BTs by recommending a practical approach to follow when designing and implementing BTs. 
Although the guidelines were based on the authors previous experiences, they lacked a broader practitioner community perspective and validation.

\looseness=-1 The empirical works of \citet{filippone2026formalisms}, \citet{dragule2025effects}, %
\citet{ghzouli2023behavior} and \citet{klauck2025surveying} investigated the modeling concepts and formalism of BTs in practice. In \citep{filippone2026formalisms}, the authors validated their findings about four behavior modeling formalism, including BTs, with domain expert by conducing questionnaire surveys. However, this work only investigated behavior models in terms of expressiveness and existing tool support for mission specification. In \citep{dragule2025effects}, the authors conducted a controlled experiment on the effectiveness and
efficiency of specifying robotics missions in BTs and state machines. They targeted undergrad students in the study, limiting  practitioner participation. \citet{ghzouli2023behavior} examined BT libraries from a software engineering design perspective and analyzed their usage in open-source projects. However, the authors did not capture practitioners' development contexts such as rationale, experiences, or design decision factors behind different choices. Finally, \citet{klauck2025surveying} conducted two surveys, one online and another in an annual developers' conference for ROS community (ROSCon 2024) following a workshop on deliberation technologies. The two surveys captured practitioners' practices regarding deliberation technologies, including behavior trees, and the use of formal verification in robotics software systems. In addition, they captured feedback for specific libraries they presented in the workshop such as \BTCPP, and they highlighted briefly gaps in available support for debugging and testing.

Compared to the contributions in this paper, previous research either 
reflects the perspective of the authors' own implementation experience rather than a broader practitioner community, or is limited in their empirical investigation aspects of BTs.
Overall, there is a clear gap between the existing body of BT research and the need for a broad understanding of practitioners real experiences and practices. 
We provide empirical evidence on the experiences, practices, and influencing factors associated with BT adoption in practice.
Our paper builds on and extends previous research and, to our knowledge, we present the first industrial technical action-research study on BTs, with a practitioner survey to expand the scope of our findings.
\section{Conclusion}

In this work, we presented an empirical mixed-method study investigating both academic and industrial practitioners' practices, experiences and influencing factors when adopting BTs in ROS-based robotic systems. We reported findings and observations across eight interest areas spanning BT design, technical integration, and the broader practical implications including team collaboration, the understandability of robotic decision-making logic and scalability.

Across our findings, practitioners valued BTs for their visual representation and communication support.
Although practitioners reported positive experiences with the most popular libraries (\BTCPP and \pytrees), they also identified challenges related to documentation, and tooling support. The limited tooling support also arose as an issue for BT scalability.

Our results showed that practitioners faced several challenging decisions, including the appropriate level of BT model granularity, the choice between single-language and mixed-language implementations of BTs, the placement of BTs within the ROS node architecture, and whether to use a planning algorithm for automatic modification of BTs.
We observed  several challenges that were prominent in the  action-research study, where participants were new to BTs, were not prominent in the survey, where respondents had more experience. 
Disentangling the two levels of experience in support, documentation, and guidelines represents a valuable direction for future research.

In the future, we would like to investigate further the source of perception discrepancy between different groups of BT practitioners for experiences and practices. Our goal would be clarifying the sources to recommend entry-level resources improvement and suggesting concrete enhancement for existing artifacts. 
Another valuable area of future work would be developing practitioner-specific guidelines for BT adoption based on our results. 
Finally, we intend to conduct an additional technical action-research study in a different industrial context to observe emerging experiences and practices. We encourage both software and robotic research communities to build upon our study and pursue similar investigations.

\begin{acknowledgements}
This work was partially supported by the Wallenberg AI, Autonomous Systems and Software Program (WASP) funded by the Knut and Alice Wallenberg Foundation.
\end{acknowledgements}

\bibliographystyle{spbasic}

\bibliography{main}

@inproceedings{wieringa2012technical,
	title={Technical action research as a validation method in information systems design science},
	author={Wieringa, Roel and Moral{\i}, Ay{\c{s}}e},
	booktitle={International Conference on Design Science Research in Information Systems},
	pages={220--238},
	year={2012},
	organization={Springer}
}

@book{wieringa2014design,
	title={Design science methodology for information systems and software engineering},
	author={Wieringa, Roel J},
	year={2014},
	publisher={Springer}
}

@book{colledanchise2018behavior,
	title="Behavior Trees in Robotics and Al: An Introduction",
	author="Colledanchise, Michele and {\"O}gren, Petter",
	year="2018",
	publisher="CRC Press"
}

@inproceedings{colledanchise2016advantages,
	title="The advantages of using behavior trees in multi robot systems",
	author="Colledanchise, Michele and Marzinotto, Alejandro and Dimarogonas, Dimos V and {\"O}gren, Petter",
	booktitle="47th International Symposium on Robotics ({ISR})",
	year="2016"
}

@article{dahl2022sequence,
	title={Sequence planner: A framework for control of intelligent automation systems},
	author={Dahl, Martin and Er{\H{o}}s, Endre and Bengtsson, Kristofer and Fabian, Martin and Falkman, Petter},
	journal={Applied Sciences},
	volume={12},
	number={11},
	pages={5433},
	year={2022},
	publisher={MDPI}
}

@article{erHos4157292framework,
	title={A Framework for Integrated Virtual Preparation and Commissioning of Intelligent Automation Systems},
	author={Er{\H{o}}s, Endre and Bengtsson, Kristofer and Dahl, Martin and Falkman, Petter},
	year={2022},
	journal={Available at SSRN 4157292}
}

@article{ghzouli2023behavior,
	title={Behavior trees and state machines in robotics applications},
	author={Ghzouli, Razan and Berger, Thorsten and Johnsen, Einar Broch and Wasowski, Andrzej and Dragule, Swaib},
	journal={IEEE Transactions on Software Engineering},
	volume={49},
	number={9},
	pages={4243--4267},
	year={2023},
	publisher={IEEE}
}

@inproceedings{colledanchise2019towards,
	title={Towards blended reactive planning and acting using behavior trees},
	author={Colledanchise, Michele and Almeida, Diogo and {\"O}gren, Petter},
	booktitle={2019 international conference on robotics and automation (ICRA)},
	pages={8839--8845},
	year={2019},
	organization={IEEE}
}

@Misc{appendix:online,
	author       = {Ghzouli, Razan},
	year         = {2026},
	title        = {{Online Appendix}},
	howpublished = {\url{https://github.com/RazanGhzouli/Behavior-Trees-for-Robotic-Systems-data}},
	note         = {Accessed: 2026-09-18}

	
}

@article{stol2018abc,
	title={The ABC of software engineering research},
	author={Stol, Klaas-Jan and Fitzgerald, Brian},
	journal={ACM Transactions on Software Engineering and Methodology (TOSEM)},
	volume={27},
	number={3},
	pages={1--51},
	year={2018},
	publisher={ACM New York, NY, USA}
}

@article{maier2017model,
	title={Model granularity in engineering design--concepts and framework},
	author={Maier, Jakob F and Eckert, Claudia M and Clarkson, P John},
	journal={Design Science},
	volume={3},
	pages={e1},
	year={2017},
	publisher={Cambridge University Press}
}

@inproceedings{rico2021optimized,
	title={Optimized Execution of PDDL Plans using Behavior Trees.},
	author={Rico, Francisco Mart{\'\i}n and Morelli, Matteo and Espinoza, Huascar and Rodr{\'\i}guez-Lera, Francisco J and Olivera, Vicente Matell{\'a}n},
	booktitle={AAMAS},
	pages={1596--1598},
	year={2021}
}

@misc{tenorth2022controlling,
	title={Controlling process of robots having a behavior tree architecture},
	author={TENORTH, Moritz},
	year={2022},
	month=may # "~24",
	publisher={Google Patents},
	note={US Patent 11,338,434}
}

@article{garcia2023software,
	title={Software variability in service robotics},
	author={Garc{\'\i}a, Sergio and Str{\"u}ber, Daniel and Brugali, Davide and Di Fava, Alessandro and Pelliccione, Patrizio and Berger, Thorsten},
	journal={Empirical Software Engineering},
	volume={28},
	number={2},
	pages={24},
	year={2023},
	publisher={Springer}
}

@inproceedings{canelas2024understanding,
	title={Understanding misconfigurations in ROS: an empirical study and current approaches},
	author={Canelas, Paulo and Schmerl, Bradley and Fonseca, Alcides and Timperley, Christopher S},
	booktitle={Proceedings of the 33rd ACM SIGSOFT International Symposium on Software Testing and Analysis},
	pages={1161--1173},
	year={2024}
}

@inproceedings{hallen2024behavior,
	title={Behavior trees in industrial applications: A case study in underground explosive charging},
	author={Hallen, Mattias and Iovino, Matteo and Sander-Tavallaey, Shiva and Smith, Christian},
	booktitle={2024 IEEE 20th International Conference on Automation Science and Engineering (CASE)},
	pages={156--162},
	year={2024},
	organization={IEEE}
}

@article{dragule2025effects,
	title={Effects of specifying robotic missions in behavior trees and state machines},
	author={Dragule, Swaib and Bainomugisha, Engineer and Pelliccione, Patrizio and Berger, Thorsten},
	journal={Journal of Computer Languages},
	pages={101330},
	year={2025},
	publisher={Elsevier}
}

@article{filippone2026formalisms,
  author={Filippone, Gianluca and Pettinari, Sara and Pelliccione, Patrizio},
  journal={IEEE Transactions on Software Engineering}, 
  title={Formalisms for Robotic Mission Specification and Execution: A Comparative Analysis}, 
  year={2026},
  volume={},
  number={},
  pages={1-32},
  doi={10.1109/TSE.2026.3725356}}

@inproceedings{schlegel2010design,
	title={Design abstraction and processes in robotics: From code-driven to model-driven engineering},
	author={Schlegel, Christian and Steck, Andreas and Brugali, Davide and Knoll, Alois},
	booktitle={International Conference on Simulation, Modeling, and Programming for Autonomous Robots},
	pages={324--335},
	year={2010},
	organization={Springer}
}

@inproceedings{ghzouli2020behavior,
	title={Behavior trees in action: a study of robotics applications},
	author={Ghzouli, Razan and Berger, Thorsten and Johnsen, Einar Broch and Dragule, Swaib and Wasowski, Andrzej},
	booktitle={Proceedings of the 13th ACM SIGPLAN international conference on software language engineering},
	pages={196--209},
	year={2020}
}

@inproceedings{petersen2009context,
	title={Context in industrial software engineering research},
	author={Petersen, Kai and Wohlin, Claes},
	booktitle={2009 3rd international symposium on empirical software engineering and measurement},
	pages={401--404},
	year={2009},
	organization={IEEE}
}

@inproceedings{kolak2020takes,
	title={It takes a village to build a robot: An empirical study of the ROS ecosystem},
	author={Kolak, Sophia and Afzal, Afsoon and Le Goues, Claire and Hilton, Michael and Timperley, Christopher Steven},
	booktitle={2020 IEEE International Conference on Software Maintenance and Evolution (ICSME)},
	pages={430--440},
	year={2020},
	organization={IEEE}
}

@inproceedings{malavolta2020you,
	title={How do you architect your robots? State of the practice and guidelines for ROS-based systems},
	author={Malavolta, Ivano and Lewis, Grace and Schmerl, Bradley and Lago, Patricia and Garlan, David},
	booktitle={Proceedings of the ACM/IEEE 42nd International Conference on Software Engineering: Software Engineering in Practice},
	pages={31--40},
	year={2020}
}

@article{creswell2003advanced,
	title={Advanced mixed methods research designs},
	author={Creswell, John W and Plano Clark, Vicki L and Gutmann, Michelle L and Hanson, William E},
	journal={Handbook of mixed methods in social and behavioral research},
	volume={209},
	number={240},
	pages={209--240},
	year={2003},
	publisher={Sage}
}

@article{storey2014r,
	title={The (r) evolution of social media in software engineering},
	author={Storey, Margaret-Anne and Singer, Leif and Cleary, Brendan and Figueira Filho, Fernando and Zagalsky, Alexey},
	journal={Future of software engineering proceedings},
	pages={100--116},
	year={2014}
}

@inproceedings{begel2010social,
	title={Social media for software engineering},
	author={Begel, Andrew and DeLine, Robert and Zimmermann, Thomas},
	booktitle={Proceedings of the FSE/SDP workshop on Future of software engineering research},
	pages={33--38},
	year={2010}
}

@article{terry2017thematic,
	title={Thematic analysis},
	author={Terry, Gareth and Hayfield, Nikki and Clarke, Victoria and Braun, Virginia and others},
	journal={The SAGE handbook of qualitative research in psychology},
	volume={2},
	number={17-37},
	pages={25},
	year={2017},
	publisher={SAGE Publications Ltd}
}

@inproceedings{garcia2020robotics,
	title={Robotics software engineering: A perspective from the service robotics domain},
	author={Garc{\'\i}a, Sergio and Str{\"u}ber, Daniel and Brugali, Davide and Berger, Thorsten and Pelliccione, Patrizio},
	booktitle={Proceedings of the 28th ACM Joint Meeting on European Software Engineering Conference and Symposium on the Foundations of Software Engineering},
	pages={593--604},
	year={2020}
}

@book{safsten2020research,
	title={Research methodology: for engineers and other problem-solvers},
	author={S{\"a}fsten, Kristina and Gustavsson, Maria},
	year={2020},
	publisher={Studentlitteratur AB}
}

@article{vears2022inductive,
	title={Inductive content analysis: A guide for beginning qualitative researchers},
	author={Vears, Danya F and Gillam, Lynn},
	journal={Focus on Health Professional Education: A Multi-Professional Journal},
	volume={23},
	number={1},
	pages={111--127},
	year={2022},
	publisher={Australian and New Zealand Association for Health Professional Educators~…}
}

@article{dortmans2022behavior,
	title={Behavior trees for smart robots practical guidelines for robot software development},
	author={Dortmans, Eric and Punter, Teade},
	journal={Journal of Robotics},
	volume={2022},
	number={1},
	pages={3314084},
	year={2022},
	publisher={Wiley Online Library}
}

@article{colledanchise2021implementation,
	title={On the implementation of behavior trees in robotics},
	author={Colledanchise, Michele and Natale, Lorenzo},
	journal={IEEE Robotics and Automation Letters},
	volume={6},
	number={3},
	pages={5929--5936},
	year={2021},
	publisher={IEEE}
}

@inproceedings{macenski2020marathon,
	title={The marathon 2: A navigation system},
	author={Macenski, Steve and Mart{\'\i}n, Francisco and White, Ruffin and Clavero, Jonatan Gin{\'e}s},
	booktitle={2020 IEEE/RSJ International Conference on Intelligent Robots and Systems (IROS)},
	pages={2718--2725},
	year={2020},
	organization={IEEE}
}

@inproceedings{martin2021plansys2,
	title={Plansys2: A planning system framework for ros2},
	author={Mart{\'\i}n, Francisco and Clavero, Jonatan Gin{\'e}s and Matell{\'a}n, Vicente and Rodr{\'\i}guez, Francisco J},
	booktitle={2021 IEEE/RSJ International Conference on Intelligent Robots and Systems (IROS)},
	pages={9742--9749},
	year={2021},
	organization={IEEE}
}

@inproceedings{bernagozzi2025model,
	title={Model-based verification and monitoring for safe and responsive robots},
	author={Bernagozzi, Stefano and Faraci, Sofia and Ghiorzi, Enrico and Pedemonte, Karim and Ferrando, Angelo and Natale, Lorenzo and Tacchella, Armando},
	booktitle={2025 IEEE International Conference on Simulation, Modeling, and Programming for Autonomous Robots (SIMPAR)},
	pages={1--6},
	year={2025},
	organization={IEEE}
}

@article{ghiorzi2024execution,
	title={Execution semantics of behavior trees in robotic applications},
	author={Ghiorzi, Enrico and Henkel, Christian and Palmas, Matteo and Klauck, Michaela and Tacchella, Armando},
	journal={arXiv preprint arXiv:2408.00090},
	year={2024}
}

@inproceedings{street2024towards,
	title={Towards a verifiable toolchain for robotics},
	author={Street, Charlie and Warsame, Yazz and Mansouri, Masoumeh and Klauck, Michaela and Henkel, Christian and Lampacrescia, Marco and Palmas, Matteo and Lange, Ralph and Ghiorzi, Enrico and Tacchella, Armando and others},
	booktitle={Proceedings of the AAAI Symposium Series},
	volume={4},
	number={1},
	pages={398--403},
	year={2024}
}

@inproceedings{quigley2009ros,
	title={ROS: an open-source Robot Operating System},
	author={Quigley, Morgan and Conley, Ken and Gerkey, Brian and Faust, Josh and Foote, Tully and Leibs, Jeremy and Wheeler, Rob and Ng, Andrew Y and others},
	booktitle={ICRA workshop on open source software},
	volume={3},
	number={3.2},
	pages={5},
	year={2009},
	organization={Kobe}
}

@article{diluoffo2018robot,
	title={Robot Operating System 2: The need for a holistic security approach to robotic architectures},
	author={DiLuoffo, Vincenzo and Michalson, William R and Sunar, Berk},
	journal={International Journal of Advanced Robotic Systems},
	volume={15},
	number={3},
	pages={1729881418770011},
	year={2018},
	publisher={SAGE Publications Sage UK: London, England}
}

@article{iovino2022survey,
	title={A survey of behavior trees in robotics and ai},
	author={Iovino, Matteo and Scukins, Edvards and Styrud, Jonathan and {\"O}gren, Petter and Smith, Christian},
	journal={Robotics and Autonomous Systems},
	volume={154},
	pages={104096},
	year={2022},
	publisher={Elsevier}
}

@article{estefo2019robot,
  title={The robot operating system: Package reuse and community dynamics},
  author={Estefo, Pablo and Simmonds, Jocelyn and Robbes, Romain and Fabry, Johan},
  journal={Journal of Systems and Software},
  volume={151},
  pages={226--242},
  year={2019},
  publisher={Elsevier}
}

@article{runeson2009guidelines,
  title={Guidelines for conducting and reporting case study research in software engineering},
  author={Runeson, Per and H{\"o}st, Martin},
  journal={Empirical software engineering},
  volume={14},
  number={2},
  pages={131--164},
  year={2009},
  publisher={Springer}
}

@article{sjoberg2007future,
	title={The Future of Empirical Methods in Software Engineering Research.},
	author={Sj{\o}berg, Dag IK and Dyb{\aa}, Tore and J{\o}rgensen, Magne},
	journal={FoSE},
	volume={7},
	number={2007},
	pages={358--378},
	year={2007}
}

@inproceedings{kampenes2008flexibility,
	title={Flexibility in research designs in empirical software engineering},
	author={Kampenes, Vigdis By and Anda, Bente and Dyb{\aa}, Tore},
	booktitle={12th International Conference on Evaluation and Assessment in Software Engineering (EASE)},
	year={2008},
	organization={BCS Learning \& Development}
}

@incollection{molleri2024teaching,
	title={Teaching Research Design in Software Engineering},
	author={Moll{\'e}ri, Jefferson Seide and Petersen, Kai},
	booktitle={Handbook on Teaching Empirical Software Engineering},
	pages={71--100},
	year={2024},
	publisher={Springer}
}

@article{moody2009physics,
  author    = {Moody, Daniel L.},
  title     = {The "Physics" of Notations: {T}oward a Scientific Basis for Constructing Visual Notations in Software Engineering},
  journal   = {IEEE Transactions on Software Engineering},
  volume    = {35},
  number    = {6},
  pages     = {756--779},
  year      = {2009},
  doi       = {10.1109/TSE.2009.67}
}

@inproceedings{izzo2024btgenbot,
  title={Btgenbot: Behavior tree generation for robotic tasks with lightweight llms},
  author={Izzo, Riccardo Andrea and Bardaro, Gianluca and Matteucci, Matteo},
  booktitle={2024 IEEE/RSJ International Conference on Intelligent Robots and Systems (IROS)},
  pages={9684--9690},
  year={2024},
  organization={IEEE}
}

@inproceedings{ao2025llm,
  title={LLM-as-BT-Planner: Leveraging LLMs for behavior tree generation in robot task planning},
  author={Ao, Jicong and Wu, Fan and Wu, Yansong and Swiki, Abdalla and Haddadin, Sami},
  booktitle={2025 IEEE International Conference on Robotics and Automation (ICRA)},
  pages={1233--1239},
  year={2025},
  organization={IEEE}
}

@inproceedings{zhou2024llm,
  title={Llm-bt: Performing robotic adaptive tasks based on large language models and behavior trees},
  author={Zhou, Haotian and Lin, Yunhan and Yan, Longwu and Zhu, Jihong and Min, Huasong},
  booktitle={2024 IEEE International Conference on Robotics and Automation (ICRA)},
  pages={16655--16661},
  year={2024},
  organization={IEEE}
}

@inproceedings{klauck2025surveying,
	title={Surveying Deliberation Practices and Methodological Needs in Robotics Software Engineering},
	author={Klauck, Michaela and Henkel, Christian and Lampacrescia, Marco and Jorgensen, Ginny},
	booktitle={Annual Conference Towards Autonomous Robotic Systems},
	pages={237--244},
	year={2025},
	organization={Springer}
}

@article{ingrand2017deliberation,
	title={Deliberation for autonomous robots: A survey},
	author={Ingrand, F{\'e}lix and Ghallab, Malik},
	journal={Artificial Intelligence},
	volume={247},
	pages={10--44},
	year={2017},
	publisher={Elsevier}
}

\end{document}